\documentclass[10pt,twocolumn,letterpaper]{article}

\usepackage[pagenumbers]{cvpr} % To force page numbers, e.g. for an arXiv version

\definecolor{cvprblue}{rgb}{0.21,0.49,0.74}
\usepackage[pagebackref,breaklinks,colorlinks,allcolors=cvprblue]{hyperref}
\usepackage[utf8]{inputenc} % allow utf-8 input
\usepackage[T1]{fontenc}    % use 8-bit T1 fonts
\usepackage{hyperref}       % hyperlinks
\usepackage{url}            % simple URL typesetting
\usepackage{booktabs}       % professional-quality tables
\usepackage{amsfonts}       % blackboard math symbols
\usepackage{nicefrac}       % compact symbols for 1/2, etc.
\usepackage{microtype}      % microtypography
\usepackage{xcolor}         % colors
\usepackage{graphicx}
\usepackage{multirow}
\usepackage{algorithm}
\usepackage{listings}
\usepackage{wrapfig}
\usepackage{arydshln}
\usepackage{rotating}

\def\paperID{9386} % *** Enter the Paper ID here
\def\confName{CVPR}
\def\confYear{2025}

\title{MVGenMaster: Scaling Multi-View Generation from Any Image via 3D Priors Enhanced Diffusion Model}

\author{%
  Chenjie Cao$^{1,2,3}$, Chaohui Yu$^{2,3}$, Shang Liu$^{2,3}$, Fan Wang$^{2}$, Xiangyang Xue$^{1}$, Yanwei Fu$^{1}$ \\
  $^1$Fudan University, 
  $^2$DAMO Academy, Alibaba Group, 
  $^3$Hupan Lab\\
  {\tt\small \{caochenjie.ccj,huakun.ych,liushang.ls,fan.w\}@alibaba-inc.com, \{xyxue,yanweifu\}@fudan.edu.cn}\\
  \small{Project page (code, model, and data): \url{https://ewrfcas.github.io/MVGenMaster}}
}

\begin{document}

\twocolumn[{
\renewcommand\twocolumn[1][]{#1}
\maketitle
\centering
\includegraphics[width=1.0\linewidth]{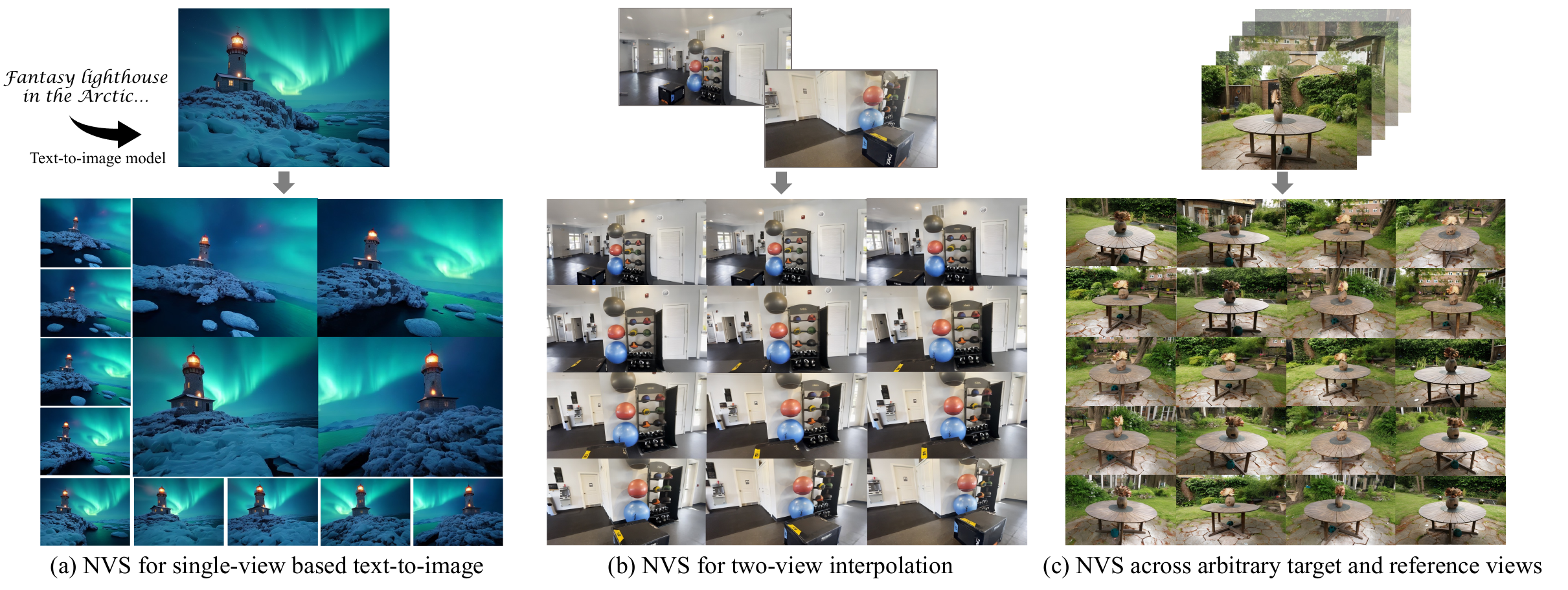}
\vspace{-0.25in}
\captionof{figure}{ 
\textbf{The proposed MVGenMaster handles various NVS scenarios properly as a master}, including (a) NVS based on single-view text-to-image conditions, (b) interpolation between two known views, and (c) flexible NVS with variable reference views and arbitrary target views. MVGenMaster enables all tasks above with a single forward process without sophisticated iterative inference and dataset updating.\label{fig:teaser}}
\vspace{0.15in}
}]

\maketitle

\begin{abstract}
We introduce \textbf{MVGenMaster}, a multi-view diffusion model enhanced with 3D priors to address versatile Novel View Synthesis (NVS) tasks. MVGenMaster leverages 3D priors that are warped using metric depth and camera poses, significantly enhancing both generalization and 3D consistency in NVS.
Our model features a simple yet effective pipeline that can generate up to 100 novel views conditioned on variable reference views and camera poses with a single forward process.
Additionally, we have developed a comprehensive large-scale multi-view image dataset called \textbf{MvD-1M}, comprising up to 1.6 million scenes, equipped with well-aligned metric depth to train MVGenMaster.
Moreover, we present several training and model modifications to strengthen the model with scaled-up datasets.
Extensive evaluations across in- and out-of-domain benchmarks demonstrate the effectiveness of our proposed method and data formulation. 
% Models and codes will be released soon.
\end{abstract}

\section{Introduction}
\label{sec:intro}
The increasing demand for high-quality 3D content is reshaping the landscape of video games, visual effects, and mixed reality devices, making it essential for real-time interactivity. 
Central to this evolution is Novel View Synthesis (NVS), a critical technology that enables the creation of 3D content by synthesizing consistent images from multiple viewpoints.  
While 3D reconstruction manners like NeRFs~\cite{mildenhall2020nerf} and 3D Gaussian Splatting (3DGS)~\cite{kerbl20233d} have achieved remarkable advancements, their reliance on dense multi-view observations limits their usability. 
Consequently, scenarios involving limited viewpoints, such as single-view NVS and two-view interpolation, remain challenging, as illustrated in \Cref{fig:teaser}.

Benefited from the developments in diffusion-based text-to-image models~\cite{rombach2022high,betker2023improving,podell2024sdxl,esser2024scaling}, NVS methods achieved significant advancements which largely alleviate the restriction of dense multi-view captures by generating novel views from reference images or textual descriptions~\cite{liu2023zero,shi2023zero123++,zhang20243d,hong2024lrm,li2024director3d,gao2024cat3d}.
Despite these advancements, intractable challenges remain in achieving high-quality NVS that is consistent, flexible, and generalizable:
1) \emph{Data limitations.} Most works rely on large-scale synthetic datasets~\cite{deitke2023objaverse,deitke2023objaversexl}, primarily targeting object-centric 3D generation~\cite{liu2023zero,shi2023MVDream,wang2023imagedream,hong2024lrm}. This focus limits their applicability to complex scene-level NVS tasks.
2) \emph{Missing 3D priors.} Currently, many diffusion-based NVS methods~\cite{sargent2024zeronvs,gao2024cat3d} are heavily reliant on 2D generations without integrating 3D priors. This restricts their ability to be scaled up while ensuring 3D consistency, particularly in out-of-domain (OOD) scenarios.
3) \emph{Lacking of flexibility.} Existing NVS techniques often lack the flexibility to accommodate arbitrary reference and target views from any viewpoint.
So they suffer from cumbersome anchor-based iterative generation~\cite{seo2024genwarp,gao2024cat3d,yu2024viewcrafter}, dataset updates~\cite{liang2024luciddreamer,yu2024wonderworld,shriram2024realmdreamer}, and test-time optimization~\cite{sargent2024zeronvs,li2024dreamscene}.
Furthermore, these methods fail to provide a straightforward solution to handle all downstream NVS requirements (\Cref{fig:teaser}) simultaneously. 

% To this end, we propose \textbf{MVGenMaster}, a novel diffusion-based \textbf{M}ulti-\textbf{V}iew \textbf{Gen}eration framework that functions as scaling multi-view generation from \textbf{Any} image.  As shown in \Cref{fig:teaser}, our model generates extensive target views with varying reference images (from a single, or two, to many more images), which enjoys both flexibility and generalization.
% Critically, our MVGenMaster is empowered via a novel 3D priors enhanced diffusion model. Thus, MVGenMaster unifies advancements from both \textit{generation} and \textit{reconstruction} through employing metric depth warping as 3D priors to ensure 3D consistency within a 2D diffusion model.

To address this, we propose \textbf{MVGenMaster}, a foundational diffusion-based framework for scaling up \textbf{M}ulti-\textbf{V}iew \textbf{Gen}eration with \textbf{any} reference and target image. 
% that scales up multi-view generation from
As shown in \Cref{fig:teaser}, MVGenMaster generates extensive target views from a flexible range of reference images with the same model, providing both versatility and generalization.  
% —whether from a single image, two images, or many more—
Critically, a key feature of MVGenMaster is the incorporation of 3D priors, allowing our model to integrate both \textit{generation} and \textit{reconstruction} with metric depth warping. This ensures consistent 3D structure within a 2D diffusion model.

Formally, our MVGenMaster builds on the foundational structure of StableDiffusion2 (SD2)~\cite{rombach2022high} and extends it for multi-view generation with cross-view attention and pose representation. Like CAT3D~\cite{gao2024cat3d}, MVGenMaster integrates full attention across all reference and target views, but with distinct input formulations. To capture dense pose presentations, we employ the Pl{\"u}cker ray~\cite{xu2023dmv3d} to denote camera poses. 
Moreover, geometric 3D priors used in MVGenMaster include the warped Canonical Coordinate Map (CCM)~\cite{li2023sweetdreamer} and RGB pixels, which rely on well-aligned metric depth and precise camera poses to maintain coherence in 3D space.

Furthermore, we collect MvD-1M to scale up MVGenMaster's training, which is a comprehensive large-scale multi-view dataset repurposed with approximately 1.6 million scenes from 12 diverse data domains. This dataset includes both object-centric and scene-level images from real-world scenarios, with dense metric depth obtained through the alignment of monocular depth estimation~\cite{yang2024depth} and Structure-from-Motion (SfM)~\cite{schoenberger2016mvs}.

Additionally, to enable MVGenMaster to synthesize multiple novel views in a single forwarding without iterative generation, we introduce an innovative, training-free key-rescaling technique. This approach addresses the challenge of attention dilution, allowing MVGenMaster to handle extremely long sequences. Moreover, we also implement several model enhancements that significantly improve MVGenMaster’s scalability trained with large-scale datasets, including noise schedule adjustment, domain switcher, multi-scale training, and exponential moving average.
% making it adaptable for large-scale generation applications.
Through these improvements, MVGenMaster consistently outperforms existing methods on both in-domain and OOD benchmarks and establishes state-of-the-art NVS results.

We highlight the key contributions of MVGenMaster as:
\begin{itemize}
    \item \textbf{Generalization.} MVGenMaster is strengthened by metric depth priors that ensure superior multi-view consistency and robust generalization across diverse scenarios.
    \item \textbf{Flexibility.} MVGenMaster is a flexible multi-view diffusion model to handle versatile downstream NVS tasks with variable target and reference views.
    % allowing the generation of arbitrary target views conditioned on various reference views.
    \item \textbf{Scalability.} We collect a large-scale multi-view dataset comprising 1.6 million scenes, specifically designed to scale up the training of MVGenMaster. Notably, all images in this dataset include metric depth for geometric warping.
\end{itemize}

\section{Related Works}
\label{sec:related_works}

\noindent\textbf{Regression-based NVS.}
Some regression-based NVS manners focused on developing promising feature encoders to produce generalized 3D representations~\cite{yu2021pixelnerf,charatan2024pixelsplat,chen2024mvsplat}, which utilize feed-forward models without per-scene optimization. 
However, these methods struggle to tackle unseen scenes and viewpoints.
In contrast, other approaches built intermediate representations, such as tri-plane, to learn NVS by volumetric or splatting rendering and regressive loss~\cite{hong2024lrm,zou2024triplane,tang2024lgm}.
Although these methods perform well in certain scenarios, they are trained from scratch without 2D priors from generative models, resulting in limited applicability, \emph{e.g.}, NVS for object-centric scenes with simple backgrounds.

\noindent\textbf{Diffusion-based NVS.}
The recent advancements in diffusion models~\cite{ho2020denoising,song2021score,song2021denoising,rombach2022high} have witnessed huge potential for image synthesis. Many pioneering studies demonstrated that diffusion models also enjoy a good capacity to address NVS tasks~\cite{chan2023generative,anciukevicius2024denoising}.
To further leverage the 2D priors from pre-trained text-to-image models~\cite{rombach2022high,betker2023improving,esser2024scaling}, 
DreamFusion~\cite{poole2023dreamfusion} proposed to use Score Distillation Sampling (SDS) to optimize 3D representations. Many follow-ups delved deeper into refining SDS techniques~\cite{haque2023instruct,kim2023collaborative,zhu2023hifa,wang2023prolificdreamer} and 3D representations~\cite{lin2023magic3d,chen2023fantasia3d,tang2024dreamgaussian}.
%Since the original text-to-image models do not adopt camera poses
Moreover, some works~\cite{liu2023zero,wu2024reconfusion,sargent2024zeronvs} incorporate camera poses into pre-trained text-to-image models to control the view generation, leading to a feed-forward NVS pipeline. This pose-conditioned NVS framework was further extended into multi-view versions~\cite{shi2023MVDream,wang2023imagedream,shi2023zero123++,liu2024syncdreamer}, which produce high-quality 3D contents.
Though diffusion-based works have been extended into scene-level NVS~\cite{sargent2024zeronvs,hollein2024viewdiff,seo2024genwarp}, they suffer from constrained 1-view generation and limited generalization.
The recent CAT3D~\cite{gao2024cat3d} serves as a foundational multi-view diffusion model, exhibiting strong performance in scene-level generation and 3D reconstruction. Our MVGenMaster takes a step forward, substantially enhancing the NVS ability with 3D priors and extensive view generation simultaneously.

\noindent\textbf{Video-based NVS.}
Video diffusion models have developed rapidly to generate high-quality videos with multiple images~\cite{ho2022video,blattmann2023stable,xing2024dynamicrafter,girdhar2024emu}. These video models can be regarded as natural NVS generators when the camera poses are controllable. So many researches focus on taming the camera trajectories of these foundational video models~\cite{guo2024animatediff,wang2024motionctrl,melas20243d,voleti2024sv3d}. Notably, recent ReconX~\cite{liu2024reconx} and ViewCrafter~\cite{yu2024viewcrafter} further integrated 3D priors from Dust3R~\cite{wang2024dust3r} to enhance the NVS capabilities.
However, these video-based NVS approaches inherit certain limitations from their base models. Firstly, they are primarily designed for sequential view synthesis, which restricts their ability to generate views from arbitrary viewpoints. Secondly, video models require temporal positional encoding, which imposes constraints on the number of generated views, resulting in less flexible NVS. 

\noindent\textbf{Depth Estimation and NVS.} 
Recent monocular depth estimation achieved great success~\cite{bhat2023zoedepth,yin2023metric3d,yang2024depth,bochkovskii2024depth}, paving the way for numerous pioneering works in NVS.
Some work iteratively warp-and-inpaint the novel views through monocular depth and 2D inpainting~\cite{fridman2023scenescape,hollein2023text2room,yu2024wonderjourney,liang2024luciddreamer,shriram2024realmdreamer,yu2024wonderworld}. While these methods show good scene-level synthesis, they are hindered by prohibitive iterative dataset updates and per-scene optimization. Moreover, most latent feature based 2D inpainting models fail to handle zoom-in camera motion properly.
Other approaches incorporate depth maps to the generative models by geometric warping~\cite{seo2024genwarp,tung2024megascenes} or prior incorporation~\cite{szymanowicz2024flash3d}. 
However, to the best of our knowledge, there is currently no research that unifies depth priors with multi-view diffusion models for NVS.

\section{Method}

\begin{figure}
\begin{center}
\includegraphics[width=1.0\linewidth]{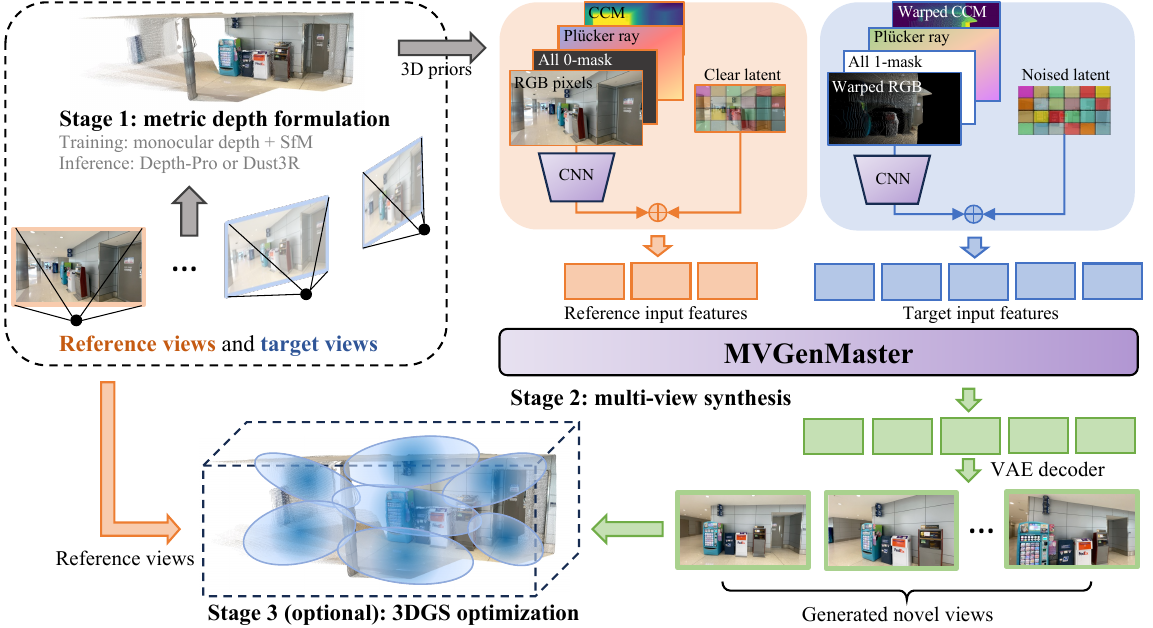}
\vspace{-0.4in}
\end{center}
   \caption{\textbf{Overall pipeline of MVGenMaster.} Inputs can be categorized into reference views (reference images and related camera poses) and target views (camera poses only). For training, we extract monocular depths from reference views and then align them with SfM to warp CCM and RGB pixels as 3D priors for target views. For inference, we utilize Depth-Pro~\cite{bochkovskii2024depth} (single-view) or Dust3R~\cite{wang2024dust3r} (multi-view) to obtain metric depth. 
   \label{fig:pipeline}}
\vspace{-0.15in}
\end{figure}

\noindent\textbf{Overview.}
The overall pipeline of MVGenMaster is summarized in \Cref{fig:pipeline}. We first introduce the training datasets and metric depth formulation for 3D priors in \Cref{sec:metric_depth}. Then the model design is illustrated in \Cref{sec:model_design}. Subsequently, the key-rescaling technique which extends MVGenMaster's synthesized view numbers is discussed in \Cref{sec:key_rescale}.

%Subsequently, MVGenMaster produces high-quality novel views to support 3D content reconstruction (3DGS).

\begin{figure}
\begin{center}
\includegraphics[width=1.0\linewidth]{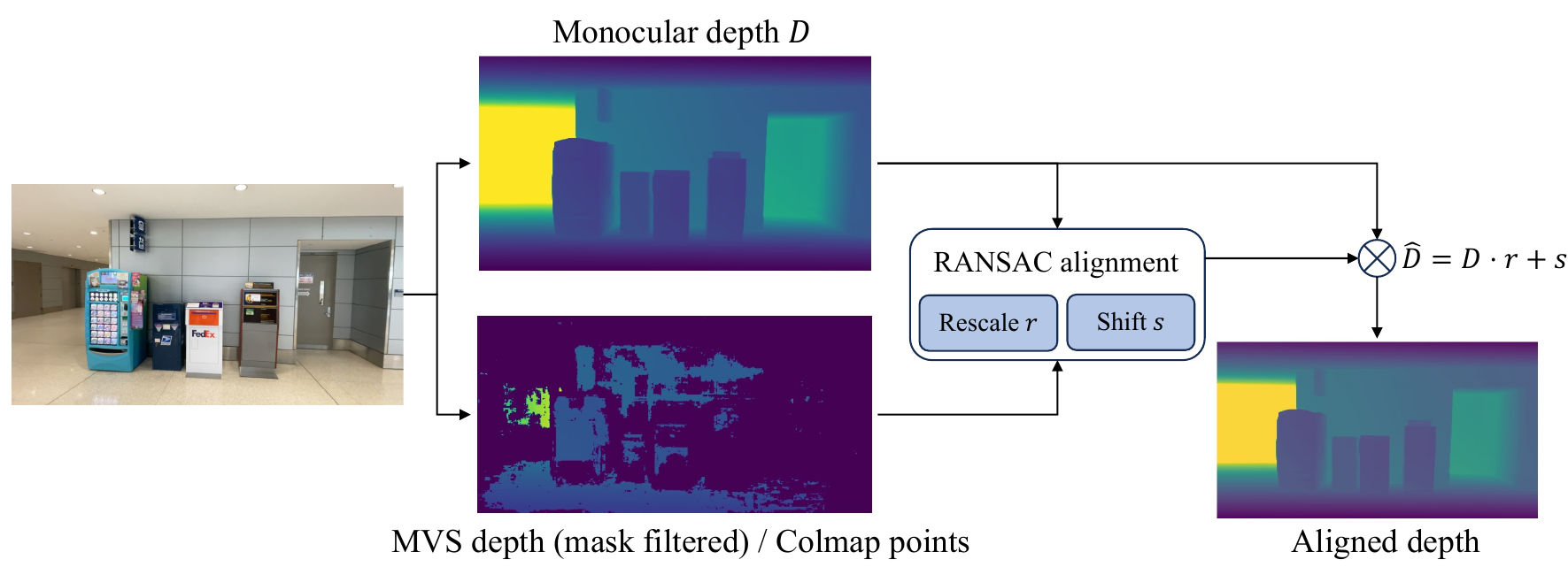}
\vspace{-0.4in}
\end{center}
   \caption{\textbf{The metric depth alignment process for the training data of MVGenMaster.} We achieve the rescale and shift coefficient by RANSAC, and then leverage them to align the monocular depth to metric one with a simple linear variation.
   \label{fig:data_processing}}
\vspace{-0.2in}
\end{figure}

\subsection{Dataset and Metric Depth Formulation}
\label{sec:metric_depth}

\noindent\textbf{Geometric Warping.}
Given relative camera viewpoint $P_{i\rightarrow j}$ and intrinsic matrices $K_i,K_j$ of view $i$ and $j$, the geometric warping function $\textrm{warp}(\cdot)$ can be presented as unprojecting pixels from view $i$ to view $j$ through the metric depth map $\hat{D}_i$. 
Thus, the relation of 2D coordinates can be denoted as:
\begin{equation}
\label{eq:warp}
x_j\simeq K_j P_{i\rightarrow j}\hat{D}_i(x_i)K^{-1}_i x_i,
\end{equation}
where $x_j$ indicates the 2D position of $x_i$ in view $j$.
Note that $\hat{D}_i$ must be the metric depth~\cite{yin2023metric3d} aligned with the extrinsic translation from SfM rather than the relative depth.

\noindent\textbf{3D Priors.}
To obtain metric depth $\hat{D}$ in training, we improve the strategy used in~\cite{tung2024megascenes}, which aligns monocular depth $D$ with sparse SfM points with RANSAC as shown in \Cref{fig:data_processing}:
\begin{equation}
\label{eq:depth_align}
\hat{D} = D \cdot r + s.
\end{equation}
To avoid collapsing to a naive solution ($r=0$), we force the minimal $r\geq 0.1\frac{std(D_{SfM})}{std(D)}$.
With metric depth maps, we formulate 3D priors as warped RGB pixels $I^{warp}$ and CCMs~\cite{li2023sweetdreamer} $C^{warp}$ from reference view $i$ to target view $j$ as:
\begin{equation}
\label{eq:depth_align}
I_{i\rightarrow j}^{warp}, C_{i\rightarrow j}^{warp} = \textrm{warp}(I_i,C_i;\hat{D}_i,P_{i\rightarrow j},K_i,K_j),
\end{equation}
where the 3D coordinate map $C_i$ is obtained by unprojecting 2D coordinates into the normalized 3D canonical space through aligned depth and camera, then encoded by Fourier embedding~\cite{vaswani2017attention}.
We integrate 3D priors warped from multiple reference views. 
In cases of conflict due to occlusion, we retain only warped $I^{warp}$ and $C^{warp}$ with minimal depth in the camera space.
Additionally, the contract function~\cite{barron2022mip} based on medium depth eliminates the infinity far depth prediction.
Notably, we warp 1:1 RGB pixels instead of 1/8 latent features for two reasons. Firstly, incorporating latent features requires additional VAE encoding~\cite{rombach2022high}, which increases training costs by 20\%. Second, using uncompressed RGB pixels enhances performance in camera magnifying scenarios, as confirmed by ablation studies.
Moreover, warped CCMs are complementary to warped RGB pixels, providing precise information about positioning and occlusion. 
This is particularly beneficial when the warped RGB pixels may be ambiguous due to significant viewpoint changes.

For the training phase, monocular depth maps are extracted by~\cite{yang2024depth}, while SfM results comprise Multi-View Stereo (MVS) depth~\cite{cao2024mvsformer++} or sparse Colmap points~\cite{schoenberger2016mvs}.
For the inference, we leverage Depth-Pro~\cite{bochkovskii2024depth} to predict both metric depth and intrinsic parameters for single-view instances, while Dust3R~\cite{wang2024dust3r} is employed to address multi-view conditions.
Instead of re-processing all training views with Dust3R, we find that inconsistent multi-view depth alignment serves as a beneficial form of regularization, which facilitates the generalization of MVGenMaster during the testing phase.
Furthermore, monocular depth alignment is much more efficient than Dust3R in processing large training scenes with more precise camera poses from SfM\footnote{Dust3R struggled to produce well-aligned point clouds for complex scenes with fixed SfM poses. Meanwhile, Dust3R failed to get camera poses (used for MVGenMaster) that are as precise as those obtained through SfM.}.

\noindent\textbf{MvD-1M Dataset.} 
We repurpose a rich collection of \textbf{M}ulti-\textbf{v}iew images with metric \textbf{D}epth, named MvD-1M.
We summarize MvD-1M in \Cref{tab:datasets}, including both object-centric and scene-level as well as indoor and outdoor scenarios.
There are 1.6 million scenes at all. Except for synthetic data, there are still 1 million diverse scenes retained with well-aligned metric depth.
Since some datasets only contain camera poses without sparse SfM points, we employ the MVS method~\cite{cao2024mvsformer++} to obtain simi-dense metric depth filtered by confidence maps.
We adopt dynamic sampling across different epochs to overcome unbalanced data distributions.
More details and visualizations are shown in supplementary.

\begin{table}
  \footnotesize
  \centering
  \setlength{\tabcolsep}{2.2pt} % 减小列间距
   \begin{tabular}{l|cccc|ccc}
\hline 
 & \multirow{2}{*}{indoor} & \multirow{2}{*}{outdoor} & \multirow{2}{*}{object} & \multirow{2}{*}{synthetic} & \multicolumn{3}{c}{scene number}\tabularnewline
\cline{6-8} \cline{7-8} \cline{8-8} 
 &  &  &  &  & {\footnotesize{}train} & {\footnotesize{}val} & {\footnotesize{}epoch}\tabularnewline
\hline 
Co3Dv2~\cite{reizenstein21co3d} & $\checkmark$ & $\checkmark$ & $\checkmark$ &  & {\footnotesize{}24.4k} & {\footnotesize{}53} & {\footnotesize{}7k}\tabularnewline
MVImgNet~\cite{yu2023mvimgnet} & $\checkmark$ & $\checkmark$ & $\checkmark$ &  & {\footnotesize{}206k} & {\footnotesize{}238} & {\footnotesize{}14.9k}\tabularnewline
DL3DV~\cite{ling2024dl3dv} & $\checkmark$ & $\checkmark$ &  &  & {\footnotesize{}9.8k} & {\footnotesize{}250} & {\footnotesize{}48k}\tabularnewline
GL3D~\cite{shen2018mirror} & $\checkmark$ & $\checkmark$ &  &  & {\footnotesize{}514} & {\footnotesize{}24} & {\footnotesize{}1.6k}\tabularnewline
Scannet++~\cite{yeshwanthliu2023scannetpp} & $\checkmark$ &  &  &  & {\footnotesize{}340} & {\footnotesize{}40} & {\footnotesize{}1.8k}\tabularnewline
3D-Front~\cite{fu20213d} & $\checkmark$ &  &  & $\checkmark$ & {\footnotesize{}10.6k} & {\footnotesize{}50} & {\footnotesize{}3.4k}\tabularnewline
Real10k~\cite{zhou2018Stereo} & $\checkmark$ & $\checkmark$ &  &  & {\footnotesize{}20.2k} & {\footnotesize{}50} & {\footnotesize{}5k}\tabularnewline
ACID~\cite{liu2021infinite} &  & $\checkmark$ &  &  & {\footnotesize{}2.6k} & {\footnotesize{}20} & {\footnotesize{}1.3k}\tabularnewline
Objaverse~\cite{deitke2023objaverse} &  &  & $\checkmark$ & $\checkmark$ & {\footnotesize{}620k} & {\footnotesize{}50} & {\footnotesize{}10k}\tabularnewline
Megascenes~\cite{tung2024megascenes} & $\checkmark$ & $\checkmark$ &  &  & {\footnotesize{}61.8k} & {\footnotesize{}100} & {\footnotesize{}30k}\tabularnewline
Aerial~\cite{googleearth} &  & $\checkmark$ &  &  & {\footnotesize{}13k} & {\footnotesize{}50} & {\footnotesize{}2k}\tabularnewline
Streetview~\cite{googleearth} & $\checkmark$ & $\checkmark$ &  &  & {\footnotesize{}677k} & {\footnotesize{}50} & {\footnotesize{}8k}\tabularnewline
\hline
Total &  &  &  &  & {\footnotesize{}1.6M} & {\footnotesize{}975} & {\footnotesize{}133k}\tabularnewline
\hline 
\end{tabular}
    \vspace{-0.1in}
    \caption{\textbf{MvD-1M dataset.} Aerial and Streetview are collected from Google Earth~\cite{googleearth}. ``object'' means object-centric. ``epoch'' indicates scene numbers dynamically sampled per-epoch. \label{tab:datasets}}
    \vspace{-0.2in}
\end{table}

\subsection{Multi-View Diffusion Model with 3D Priors}
\label{sec:model_design}

The multi-view synthesis is accomplished by the Latent Diffusion Model (LDM) in MVGenMaster as shown in stage 2 of \Cref{fig:pipeline}. 
Given a set of reference images $I_{r}$, along with Pl{\"u}cker ray embedding~\cite{xu2023dmv3d} for both reference and target camera poses ($\mathbf{P}_{r},\mathbf{P}_{t}$), and 3D priors mentioned above, the LDM $\theta$ learns to synthesize images of target views as:
\begin{equation}
\label{eq:mv_diff}
p_\theta(z_{t}|z_{r},\mathbf{P}_{r},C_{r},I_{r},\mathbf{P}_{t},C^{warp}_{r\rightarrow t},I^{warp}_{r\rightarrow t},\mathbf{M}),
\end{equation}
where $z_r$ and $z_t$ represent latent features of reference and target views; $\mathbf{M}$ is a binary mask that distinguishes reference and target features.
Unlike CAT3D~\cite{gao2024cat3d}, we incorporate Pl{\"u}cker ray embedding to confirm dense camera presentations. 
All the aforementioned features form the input for MVGenMaster. RGB pixels, Pl{\"u}cker ray, and masks are maintained in 1:1 scale, while CCMs and latent features are downscaled to 1/8 scale. 
Note that we only use VAE to get latent features, while other inputs are encoded by a lightweight CNN trained from scratch and added to latent features. 
Ablation studies in \Cref{tab:ablation1} show that this straightforward incorporation of 3D priors results in significant improvements.
The backbone of MVGenMaster is built upon the v-prediction based SD2~\cite{rombach2022high}, following the design of CAT3D with 3D full attention across all views as well as discarding the textual branch and cross-attention modules.
MVGenMaster is trained with $N+M=8$, where $N,M$ mean reference and target view numbers respectively. During the first training phase, we set $N=3$ as CAT3D. In the second phase, the reference views are dynamically adjusted to $N=1\sim3$. MVGenMaster can be generalized to varying $N$ and $M$ by key-rescaling, as outlined in \Cref{sec:key_rescale}.

\noindent\textbf{Details of Model Modification.}
% \paragraph{Details of Model Modification.}
As discussed in~\cite{shi2023zero123++,gao2024cat3d}, we find a noisier diffusion schedule is critical to train high-resolution multi-view diffusion models. 
When multi-view images of the same scene are corrupted by noise, they maintain a higher Signal-to-Noise Ratio (SNR) compared to the single-view image, which over-simplifies model training by leaking information.
Thus we replace the scaled linear schedule with the noisier linear schedule and adopt ZeroSNR~\cite{lin2024common} to train MVGenMaster.
Additionally, we find that applying qk-norm~\cite{henry-etal-2020-query} before self-attention modules prominently facilitates training stability.
Moreover, some datasets used in MVGenMaster are very different from others. For example, images from Megascenes~\cite{tung2024megascenes} were not captured at the same time, suffering from disparate illuminations and appearances.
In contrast, Objaverse~\cite{deitke2023objaverse} is an object-centric rendered dataset, while its background is overly simplistic compared to other datasets.
To unify all these diverse datasets together, we re-purpose the domain switcher used in Wonder3D~\cite{long2024wonder3d}. 
Different from Wonder3D which uses class labels to control various generation domains, we assign different class labels to Megascenes, Objaverse, and other regular multi-view datasets to eliminate the domain gap during the joint training. 
The encoded class embedding is added to the timestep embedding of the diffusion model, while we set the class label to the regular multi-view dataset for consistent appearance in the inference phase.
Furthermore, we find that the multi-scale training and Exponential Moving Average (EMA), commonly utilized in large model training, substantially enhance the performance of NVS.
% including domain switcher, qk-norm, multi-scale training, and Exponential Moving Average (EMA),

\subsection{Key-Rescaling for View Extension}
\label{sec:key_rescale}

\begin{figure}
\begin{center}
\includegraphics[width=1.0\linewidth]{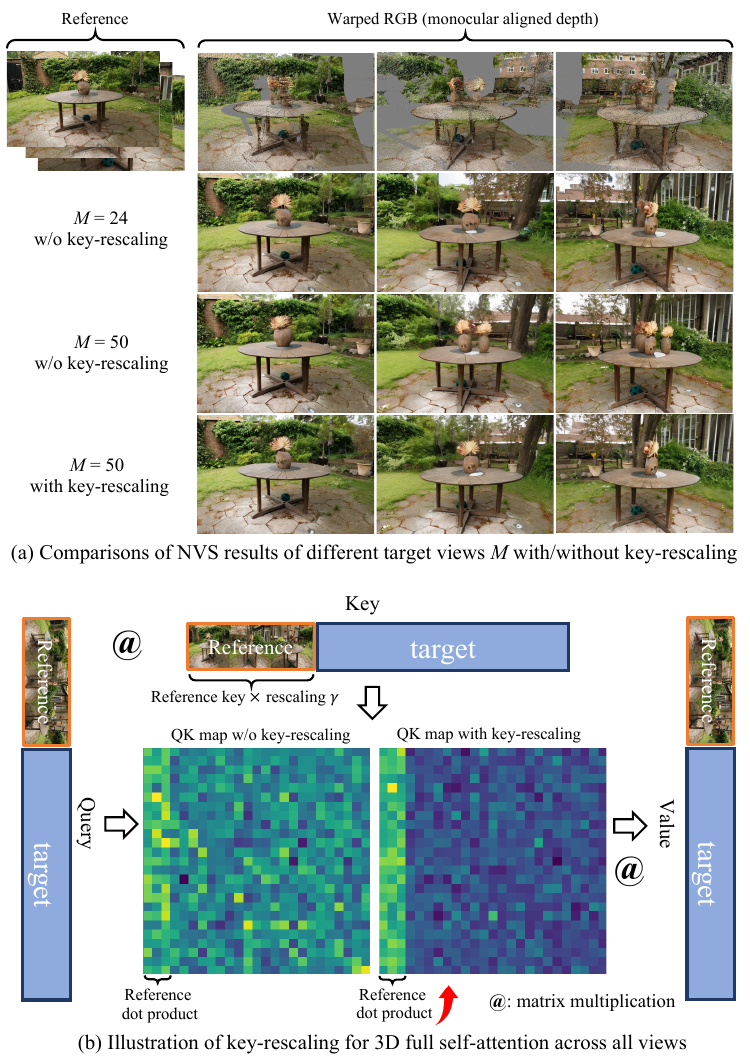}
\vspace{-0.35in}
\end{center}
   \caption{\textbf{Key-rescaling.} We employ 3-view references with ambiguous 3D priors as noisy conditions for long sequential generation. Key-rescaling enhances the reference guidance and eliminates attention dilution, resulting in better NVS with mass target views.
   \label{fig:key_rescale}}
\vspace{-0.25in}
\end{figure}

\noindent\textbf{Analysis.} Generating novel views for arbitrary viewpoints and view numbers is important in adapting models for various downstream NVS tasks.
Unlike anchor-based iterative generation~\cite{gao2024cat3d,yu2024viewcrafter}, our approach produces all views simultaneously, ensuring superior consistency and quality without the artifact accumulations as verified in the ablation study of \Cref{tab:abaltion_view_num}.
Besides, different from NVS based on video priors~\cite{voleti2024sv3d,melas20243d,liu2024reconx,yu2024viewcrafter}, as our method can handle unordered NVS from any set of viewpoints. Additionally, since MVGenMaster operates without temporal positional encoding, it can extend views without being hindered by unseen temporal positions. 
However, we discovered that naively extending MVGenMaster for extremely long sequences is non-trivial. As shown in \Cref{tab:abaltion_view_num}, the performance suffers from significant degradation, when the target view number exceeds 25.
To investigate the cause of degradation, we analyze the qualitative result of a toy case in \Cref{fig:key_rescale}(a).
In this instance, we intentionally incorporate ambiguous 3D priors warped from monocular aligned depth. 
Our model performed adequately with $M=24$ but failed significantly when increasing $M$ to 50. 
A notable issue arose where multiple vases were inaccurately synthesized in the novel views at $M=50$, which is like overfitting to the ambiguous 3D priors of responding views.
This overfitting issue is fundamentally caused by \textit{attention dilution}, occurring in extremely long sequential cases that have not been adapted during training.
Particularly, the attention dilution undermines the reference guidance, because less attention is assigned to reference views, resulting in inferior appearance and structure.
Thus, the model tends to over-depend on ``unreliable'' 3D priors from each view.

\noindent\textbf{Solution.}
We propose the key-rescaling, a training-free technique to improve the reference guidance in self-attention modules as shown in \Cref{fig:key_rescale}(b). Formally, MVGenMaster adopts 3D full self-attention across all views, while key-rescaling multiplies a constant $\gamma$ to the reference views of key features in all self-attention modules.
Therefore, all reference views would achieve higher dot product scores before the softmax operation, thereby allowing the aggregation to focus more on reference guidance and balance the attention dilution. Empirically, we find that $\gamma=1.2$ performs well in most scenarios when 158 $\geq (N+M) >$ 28 (158-view NVS is the upper bound of an 80G GPU). More detailed ablation studies are discussed in the supplementary.
As verified in \Cref{fig:key_rescale} and our ablation studies, MVGenMaster strengthened by key-rescaling exhibits strong generalization across various target views.
All experiments of MVGenMaster in this paper are accomplished by a single forward process.
Moreover, the proposed key-rescaling can be seamlessly operated to the key features before the attention module, making it compatible with the widely used FlashAttention2~\cite{dao2024flashattention}.

% \subsection{3D Reconstruction}
% \label{sec:3dgs}

\section{Experiments}

\noindent\textbf{Implementation Details.}
MVGenMaster is trained on 12 datasets as summarized in \Cref{tab:datasets}. To verify the generalization of MVGenMaster, we further adopt the validation set from DTU~\cite{aanaes2016large}, MipNeRF-360~\cite{barron2022mip}, Tanks-and-Temples~\cite{Knapitsch2017}, and ETH3D~\cite{schops2017multi} as the zero-shot benchmark.
We train MVGenMaster with view number 8 and batch size 512 on 16 A800 GPUs for 600k steps. For the first 350k steps, MVGenMaster is trained with $N=3$ reference views; the learning rate is 5e-5. Then, we reduced the learning rate to 2.5e-5, and further fine-tuned MVGenMaster with $N=1\sim3$ reference views. We set the EMA decay to 0.9995, and set the dropout rates of 15\% and 10\% for 3D priors and Pl{\"u}cker ray respectively as the classifier-free guidance (CFG) training. For the inference, we set the CFG scale to 2. Moreover, we employ the multi-scale training to MVGenMaster, while the training images are resized and cropped from 320$\times$768 to 768$\times$320. More details are discussed in the supplementary.

\subsection{Results of MVGenMaster}

\begin{table}  \small
  \centering
  \setlength{\tabcolsep}{2pt} % 减小列间距
  \begin{tabular}{l|ccc|ccc}
    \hline 
    \textbf{\footnotesize{}Datasets} & \multicolumn{3}{c|}{\textbf{\footnotesize{}Ordered}} & \multicolumn{3}{c}{\textbf{\footnotesize{}Unordered}}\tabularnewline
    \cline{2-7} \cline{3-7} \cline{4-7} \cline{5-7} \cline{6-7} \cline{7-7} 
    {\footnotesize{}Methods} & {\footnotesize{}PSNR$\uparrow$} & {\footnotesize{}SSIM$\uparrow$} & {\footnotesize{}LPIPS$\downarrow$} & {\footnotesize{}PSNR$\uparrow$} & {\footnotesize{}SSIM$\uparrow$} & {\footnotesize{}LPIPS$\downarrow$}\tabularnewline
    \hline 
    \textbf{\footnotesize{}CO3D+MVImgNet} &  &  &  &  &  & \tabularnewline
    % {\footnotesize{}ZeroNVS} &  &  &  &  &  & \tabularnewline
    {\footnotesize{}ViewCrafter~\cite{yu2024viewcrafter}} & {\footnotesize{}15.347} & {\footnotesize{}0.504} & {\footnotesize{}0.467} & {\footnotesize{}14.139} & {\footnotesize{}0.328} & {\footnotesize{}0.558}\tabularnewline
    {\footnotesize{}CAT3D{*}~\cite{gao2024cat3d}} & {\footnotesize{}17.296} & {\footnotesize{}0.509} & {\footnotesize{}0.397} & {\footnotesize{}17.540} & {\footnotesize{}0.512} & {\footnotesize{}0.379}\tabularnewline
    {\footnotesize{}MVGenMaster (1-view)} & {\footnotesize{}18.484} & {\footnotesize{}0.569} & {\footnotesize{}0.325} & {\footnotesize{}18.619} & {\footnotesize{}0.573} & {\footnotesize{}0.316}\tabularnewline
    {\footnotesize{}MVGenMaster (3-view)} & \textbf{\footnotesize{}18.964} & \textbf{\footnotesize{}0.583} & \textbf{\footnotesize{}0.306} & \textbf{\footnotesize{}21.466} & \textbf{\footnotesize{}0.653} & \textbf{\footnotesize{}0.220}\tabularnewline
    \hline 
    \textbf{\footnotesize{}DL3DV+Real10k} &  &  &  &  &  & \tabularnewline
    % {\footnotesize{}ZeroNVS} &  &  &  &  &  & \tabularnewline
    {\footnotesize{}ViewCrafter~\cite{yu2024viewcrafter}} & {\footnotesize{}13.279} & {\footnotesize{}0.352} & {\footnotesize{}0.528} & {\footnotesize{}12.711} & {\footnotesize{}0.328} & {\footnotesize{}0.558}\tabularnewline
    {\footnotesize{}CAT3D{*}~\cite{gao2024cat3d}} & {\footnotesize{}13.650} & {\footnotesize{}0.366} & {\footnotesize{}0.488} & {\footnotesize{}14.006} & {\footnotesize{}0.378} & {\footnotesize{}0.473}\tabularnewline
    {\footnotesize{}MVGenMaster (1-view)} & {\footnotesize{}15.476} & {\footnotesize{}0.458} & {\footnotesize{}0.381} & {\footnotesize{}15.729} & {\footnotesize{}0.468} & {\footnotesize{}0.376}\tabularnewline
    {\footnotesize{}MVGenMaster (3-view)} & \textbf{\footnotesize{}16.177} & \textbf{\footnotesize{}0.478} & \textbf{\footnotesize{}0.352} & \textbf{\footnotesize{}18.296} & \textbf{\footnotesize{}0.552} & \textbf{\footnotesize{}0.266}\tabularnewline
    \hline 
    \textbf{\footnotesize{}Zero-shot Datasets} &  &  &  &  &  & \tabularnewline
    % {\footnotesize{}ZeroNVS} &  &  &  &  &  & \tabularnewline
    {\footnotesize{}ViewCrafter~\cite{yu2024viewcrafter}} & {\footnotesize{}11.431} & {\footnotesize{}0.338} & {\footnotesize{}0.603} & {\footnotesize{}11.039} & {\footnotesize{}0.318} & {\footnotesize{}0.630}\tabularnewline
    {\footnotesize{}CAT3D{*}~\cite{gao2024cat3d}} & {\footnotesize{}10.865} & {\footnotesize{}0.301} & {\footnotesize{}0.652} & {\footnotesize{}10.878} & {\footnotesize{}0.296} & {\footnotesize{}0.642}\tabularnewline
    {\footnotesize{}MVGenMaster (1-view)} & {\footnotesize{}12.593} & {\footnotesize{}0.383} & {\footnotesize{}0.486} & {\footnotesize{}12.879} & {\footnotesize{}0.394} & {\footnotesize{}0.466}\tabularnewline
    {\footnotesize{}MVGenMaster (3-view)} & \textbf{\footnotesize{}13.718} & \textbf{\footnotesize{}0.418} & \textbf{\footnotesize{}0.417} & \textbf{\footnotesize{}15.533} & \textbf{\footnotesize{}0.491} & \textbf{\footnotesize{}0.319}\tabularnewline
    \hline 
    \end{tabular}
  \vspace{-0.1in}
  \caption{\textbf{Quantitative results of NVS.} CAT3D* is re-implemented as MVGenMaster without 3D priors. `Ordered': viewpoints are sorted with specific trajectories. `Unordered': random viewpoints.\label{tab:nvs_results}}
  \vspace{-0.2in}
\end{table}
% 1-V/3-V: 1/3 views.

\begin{figure*}
\begin{center}
\includegraphics[width=0.88\linewidth]{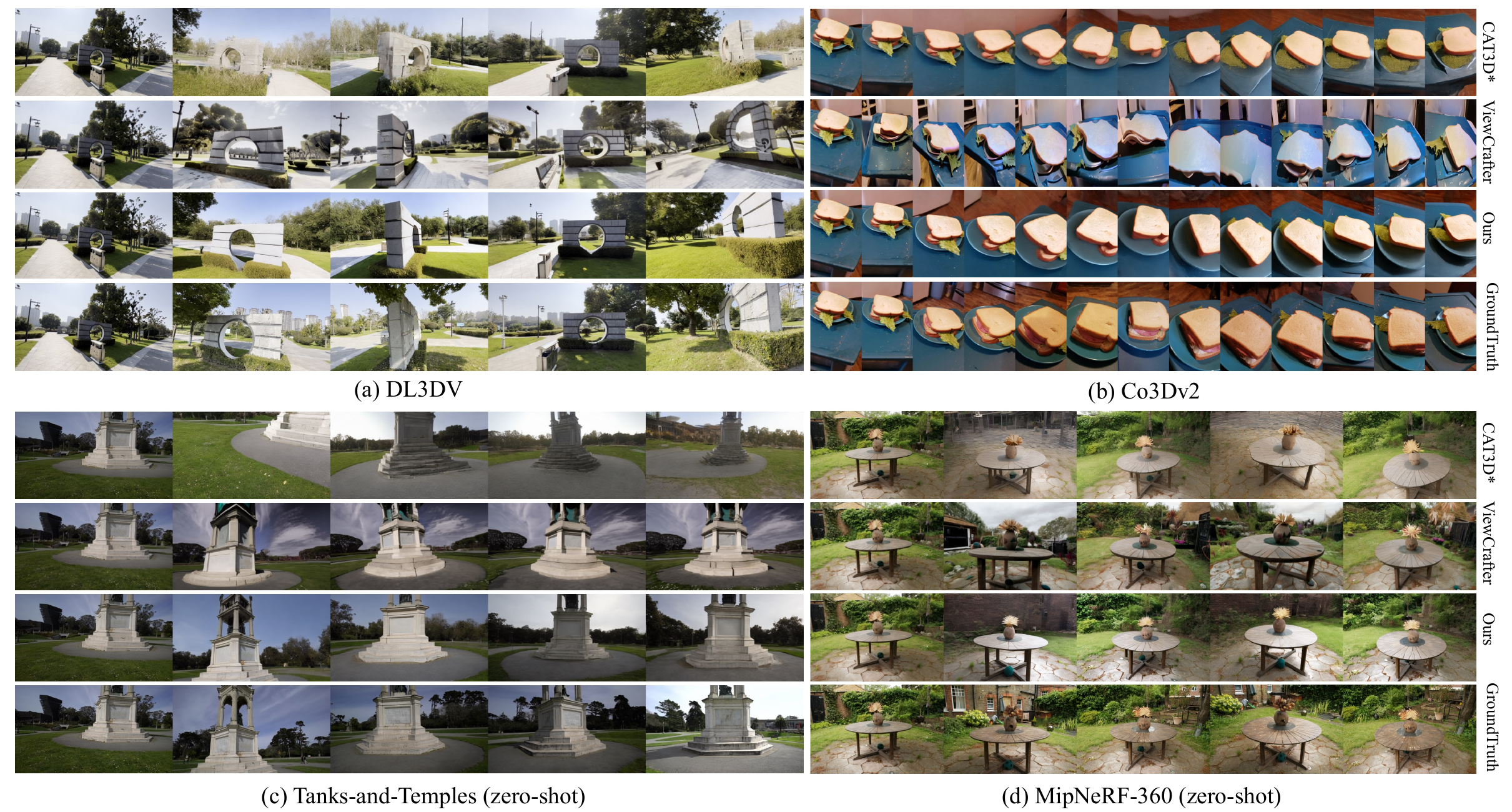}
\vspace{-0.25in}
\end{center}
   \caption{\textbf{Qualitative NVS results compared among CAT3D*, ViewCrafter, and our MVGenMaster.} The synthesis is based on ($N=1$) reference view and ($M=24$) target views.
   The leftmost column displays the reference view, while the remaining visualizations are uniformly sampled from the 24-frame generation due to page limitation. 
   \label{fig:nvs_qualitative}}
\vspace{-0.15in}
\end{figure*}

\noindent\textbf{Quantitative Results.}
We comprehensively compare the quantitative NVS results in \Cref{tab:nvs_results}, including object-centric NVS~\cite{reizenstein21co3d,yu2023mvimgnet}, scene-level NVS~\cite{ling2024dl3dv,zhou2018Stereo}, and the aforementioned zero-shot benchmark~\cite{aanaes2016large,barron2022mip,Knapitsch2017,schops2017multi}.
The competitors comprise ViewCrafter~\cite{yu2024viewcrafter}, CAT3D~\cite{gao2024cat3d}, and the proposed MVGenMaster. Since CAT3D was not open-released, we re-implemented it following the setting of MVGenMaster without 3D priors, denoted as CAT3D*. 
Furthermore, because ViewCrafter is restricted to synthesizing 25 frames at once, we set the total view number $N+M=25$ to all methods in \Cref{tab:nvs_results} for fair comparisons.
MVGenMaster outperforms other competitors under the 1-view condition in both object-centric and scene-level scenarios, while 3D priors enjoy substantial improvements relative to CAT3D*.
Though ViewCrafter slightly performs better than CAT3D* on the zero-shot test set due to the point clouds from Dust3R~\cite{wang2024dust3r}, our method still enjoys a considerable lead overall. 
Moreover, MVGenMaster can properly handle NVS with unordered viewpoints, with further improvements evident when additional reference views are available (3-view). This demonstrates the flexibility of the proposed approach.

\noindent\textbf{Qualitative Results.}
We show the qualitative results of NVS in \Cref{fig:nvs_qualitative}, with all outcomes derived from a single-view reference and 24-view generation.
For the scene-level result, ViewCrafter suffers from intractable deformation with large viewpoint changes. The re-implemented CAT3D* struggles to maintain accurate 3D structures without 3D priors. Moreover, ViewCrafter lacks generalization capabilities to object-centric scenarios, while CAT3D* hallucinates some object's parts. 
In contrast, MVGenMaster consistently delivers superior results across both scene-level and object-centric scenarios, excelling in both structure and appearance.
Furthermore, MVGenMaster also performs well in the challenging 1-view NVS on zero-shot datasets, producing stable and cohesive backgrounds alongside visually impressive foregrounds.
More results are shown in the supplementary.

\begin{figure*}
\begin{center}
\includegraphics[width=0.8\linewidth]{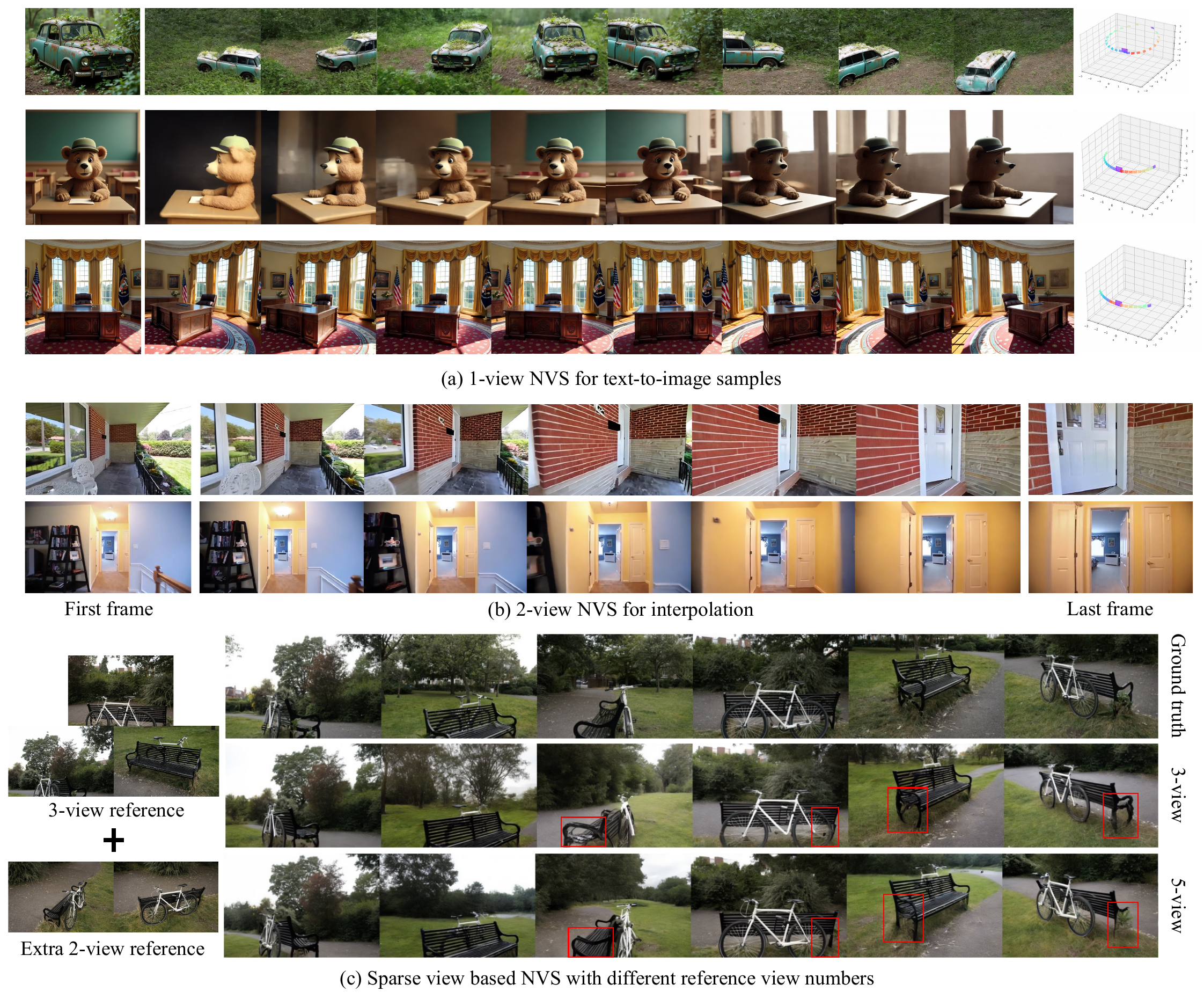}
\vspace{-0.3in}
\end{center}
   \caption{\textbf{MVGenMaster handles different NVS downstream tasks with a flexible pipeline.} 
   \label{fig:nvs_application}}
\vspace{-0.15in}
\end{figure*}

\noindent\textbf{Generalization for Various NVS Tasks.}
Thanks to the flexible design of MVGenMaster, our model can be generalized to various NVS tasks as shown in \Cref{fig:nvs_application}.
With 1-view conditions, we confirm that MVGenMaster could tackle text-to-image\footnote{\url{https://github.com/black-forest-labs/flux}} NVS, utilizing monocular metric depth and focal length predicted by Depth-Pro~\cite{bochkovskii2024depth}.
Besides, given the first and the last frames, MVGenMaster seamlessly interpolates the intermediate frames between two camera poses estimated by Dust3R~\cite{wang2024dust3r}. 
Moreover, the performance of MVGenMaster can be further enhanced by incorporating more reference views as verified in \Cref{fig:nvs_application}(c).

% \subsection{Results of Reconstruction}

\begin{table*}
  \small
  \centering
    \begin{tabular}{l|ccc|ccc|ccc}
    \hline 
    \textbf{\small{}Condition views (all)} & \multicolumn{3}{c|}{{\small{}Tanks-and-Temples}} & \multicolumn{3}{c|}{{\small{}DTU}} & \multicolumn{3}{c}{{\small{}MipNeRF-360}}\tabularnewline
    \cline{2-10} \cline{3-10} \cline{4-10} \cline{5-10} \cline{6-10} \cline{7-10} \cline{8-10} \cline{9-10} \cline{10-10} 
    {\small{}Methods} & {\small{}PSNR$\uparrow$} & {\small{}SSIM$\uparrow$} & {\small{}LPIPS$\downarrow$} & {\small{}PSNR$\uparrow$} & {\small{}SSIM$\uparrow$} & {\small{}LPIPS$\downarrow$} & {\small{}PSNR$\uparrow$} & {\small{}SSIM$\uparrow$} & {\small{}LPIPS$\downarrow$}\tabularnewline
    \hline 
    \textbf{\small{}2-view (25)} &  &  &  &  &  &  &  &  & \tabularnewline
    {\footnotesize{}ViewCrafter-sparse~\cite{yu2024viewcrafter}} & 13.408 & 0.416 & 0.462 & 12.411 & 0.406 & 0.525 & 12.966 & 0.233 & 0.600\tabularnewline
    {\footnotesize{}CAT3D{*}~\cite{gao2024cat3d}} & 12.525 & 0.375 & 0.531 & 11.756 & 0.354 & 0.618 & 12.900 & 0.211 & 0.618\tabularnewline
    {\footnotesize{}MVGenMaster} & \textbf{14.790} & \textbf{0.491} & \textbf{0.332} & \textbf{15.574} & \textbf{0.536} & \textbf{0.325} & \textbf{13.836} & \textbf{0.287} & \textbf{0.498}\tabularnewline
    \hline 
    \textbf{3-view (100)} &  &  &  &  &  &  &  &  & \tabularnewline
    {\footnotesize{}MVSplat~\cite{chen2024mvsplat}} & 8.602 & 0.190 & 0.649 & 10.772 & 0.271 & 0.557 & 11.379 & 0.171 & 0.691\tabularnewline
    {\footnotesize{}CAT3D{*}~\cite{gao2024cat3d}} & 11.758 & 0.351 & 0.745 & 11.268 & 0.365 & 0.662 & 13.609 & 0.263 & 0.714\tabularnewline
    {\footnotesize{}MVGenMaster} & \textbf{14.669} & \textbf{0.473} & \textbf{0.473} & \textbf{15.856} & \textbf{0.585} & \textbf{0.314} & \textbf{15.543} & \textbf{0.356} & \textbf{0.539}\tabularnewline
    \hline 
    \end{tabular}
    \vspace{-0.1in}
    \caption{\textbf{Quantitative results of reconstruction.} CAT3D* is re-implemented following the setting of MVGenMaster without 3D priors.\label{tab:recon}}
    \vspace{-0.1in}
\end{table*}

\noindent\textbf{Reconstruction.}
We evaluate the performance of 3DGS reconstruction~\cite{kerbl20233d} based on NVS results in \Cref{tab:recon}.
% The LPIPS loss~\cite{} is incorporated while other implementations follow the vanilla 3DGS~\cite{}.
For the 2-view scenario, the first and last frames are provided, while NVS methods produce other 23 frames; there are 25 frames at all. We compare ViewCafter-sparse, CAT3D*, and our method here, where ViewCrafter-sparse has been fine-tuned with the first and last frames. Our MVGenMaster enjoys preferable performance on zero-shot datasets.
For the 3-view scenario, CAT3D* and MVGenMaster generate 100 frames to support the 3DGS.
Especially, MVSplat~\cite{chen2024mvsplat} fails when there is insufficient view overlap.
Notably, MVGenMaster achieves prominent advancements compared to other methods benefiting from good consistency achieved through key-rescaling.
We claim that results from \Cref{tab:recon} are under a challenging but efficient setting, \ie, 3DGS based on once NVS inference, without cumbersome iterative generation~\cite{gao2024cat3d} and distillation~\cite{wu2024reconfusion}. We further compare our method to other closed-source methods in the supplementary.

\begin{table*}
  \footnotesize
  \centering
  \setlength{\tabcolsep}{2pt}
    \begin{tabular}{cccccccccccc|ccc}
    \hline 
    Co3Dv2 & MVImgNet & DL3DV & GL3D & Scannet++ & 3D-Front & Real10k & ACID & Objaverse & Megascenes & Aerial & Streetview & PSNR{\small{}$\uparrow$} & SSIM{\small{}$\uparrow$} & LPIPS{\small{}$\downarrow$}\tabularnewline
    \hline 
    $\checkmark$ & $\checkmark$ & $\checkmark$ & $\checkmark$ & $\checkmark$ & $\checkmark$ &  &  &  &  &  &  & 14.869 & 0.472 & 0.354\tabularnewline
    $\checkmark$ & $\checkmark$ & $\checkmark$ & $\checkmark$ & $\checkmark$ & $\checkmark$ & $\checkmark$ & $\checkmark$ &  &  &  &  & 15.126 & 0.477 & 0.351\tabularnewline
    $\checkmark$ & $\checkmark$ & $\checkmark$ & $\checkmark$ & $\checkmark$ & $\checkmark$ & $\checkmark$ & $\checkmark$ & $\checkmark$ & $\checkmark$ &  &  & 15.081 & 0.496 & 0.345\tabularnewline
    $\checkmark$ & $\checkmark$ & $\checkmark$ & $\checkmark$ & $\checkmark$ & $\checkmark$ & $\checkmark$ & $\checkmark$ & $\checkmark$ & $\checkmark$ & $\checkmark$ & $\checkmark$ & \textbf{15.641} & \textbf{0.503} & \textbf{0.326}\tabularnewline
    \hline 
    \end{tabular}
    \vspace{-0.125in}
    \caption{\textbf{Dataset scalability of MVGenMaster.} Results are evaluated on the zero-shot benchmark.\label{tab:dataset_scalability}}
    \vspace{-0.2in}
\end{table*}

\begin{table}
  \small
  \centering
  \renewcommand{\arraystretch}{0.8}
  \setlength{\tabcolsep}{1.5pt} % 减小列间距
    \begin{tabular}{c|c|c|c|c|c|c|ccc}
    \hline 
    \multirow{2}{*}{{\footnotesize{}cam}} & \multirow{2}{*}{{\footnotesize{}qk-norm}} & \multirow{2}{*}{{\footnotesize{}CCM}} & \multicolumn{2}{c|}{{\footnotesize{}prior type}} & \multicolumn{2}{c|}{{\footnotesize{}warp}} & \multirow{2}{*}{{\footnotesize{}PSNR$\uparrow$}} & \multirow{2}{*}{{\footnotesize{}SSIM$\uparrow$}} & \multirow{2}{*}{{\footnotesize{}LPIPS$\downarrow$}}\tabularnewline
    \cline{4-7} \cline{5-7} \cline{6-7} \cline{7-7} 
     &  &  & {\footnotesize{}conv} & {\footnotesize{}cross-attn} & {\footnotesize{}latent} & {\footnotesize{}pixel} &  &  & \tabularnewline
    \hline 
    {\footnotesize{}$\checkmark$} &  &  &  &  &  &  & {\footnotesize{}15.348} & {\footnotesize{}0.479} & {\footnotesize{}0.462}\tabularnewline
    {\footnotesize{}$\checkmark$} & {\footnotesize{}$\checkmark$} &  &  &  &  &  & {\footnotesize{}15.521} & {\footnotesize{}0.483} & {\footnotesize{}0.451}\tabularnewline
    {\footnotesize{}$\checkmark$} & {\footnotesize{}$\checkmark$} & {\footnotesize{}$\checkmark$} & {\footnotesize{}$\checkmark$} &  &  &  & {\footnotesize{}17.168} & {\footnotesize{}0.532} & {\footnotesize{}0.360}\tabularnewline
    {\footnotesize{}$\checkmark$} & {\footnotesize{}$\checkmark$} & {\footnotesize{}$\checkmark$} &  & {\footnotesize{}$\checkmark$} &  &  & {\footnotesize{}15.830} & {\footnotesize{}0.498} & {\footnotesize{}0.436}\tabularnewline
    {\footnotesize{}$\checkmark$} & {\footnotesize{}$\checkmark$} & {\footnotesize{}$\checkmark$} & {\footnotesize{}$\checkmark$} &  & {\footnotesize{}$\checkmark$} &  & {\footnotesize{}17.521} & {\footnotesize{}0.550} & \textbf{\footnotesize{}0.345}\tabularnewline
    {\footnotesize{}$\checkmark$} & {\footnotesize{}$\checkmark$} & {\footnotesize{}$\checkmark$} & {\footnotesize{}$\checkmark$} &  &  & {\footnotesize{}$\checkmark$} & \textbf{\footnotesize{}17.651} & \textbf{\footnotesize{}0.554} & {\footnotesize{}0.346}\tabularnewline
     & {\footnotesize{}$\checkmark$} & {\footnotesize{}$\checkmark$} & {\footnotesize{}$\checkmark$} &  &  & {\footnotesize{}$\checkmark$} & {\footnotesize{}17.514} & {\footnotesize{}0.553} & {\footnotesize{}0.352}\tabularnewline
    \hline 
    \end{tabular}
    \vspace{-0.125in}
      \caption{\textbf{Ablation studies on our in-domain validation set.} Models are trained on our training subset.\label{tab:ablation1}}
  \vspace{-0.1in}
\end{table}

\begin{table}
  \small
  \centering
  \setlength{\tabcolsep}{3.5pt} % 减小列间距
    \begin{tabular}{ccc|ccc}
    \hline 
    domain switcher & multi-scale & EMA & PSNR{\small{}$\uparrow$} & SSIM{\small{}$\uparrow$} & LPIPS{\small{}$\downarrow$}\tabularnewline
    \hline 
     &  &  & 14.286 & 0.462 & 0.384\tabularnewline
    $\checkmark$ &  &  & 14.956 & 0.472 & 0.364\tabularnewline
    $\checkmark$ & $\checkmark$ &  & 15.211 & 0.482 & 0.339\tabularnewline
    $\checkmark$ & $\checkmark$ & $\checkmark$ & \textbf{15.641} & \textbf{0.503} & \textbf{0.326}\tabularnewline
    \hline 
    \end{tabular}
    \vspace{-0.125in}
    \caption{\textbf{Ablation studies of scaling up MVGenMaster with the full training set.} Results are based on the zero-shot test set.\label{tab:ablation_scaling_up}}
  \vspace{-0.1in}
\end{table}

\begin{table}
  \small
  \centering
    \setlength{\tabcolsep}{3.5pt}
    \begin{tabular}{ccc|ccc}
    \hline 
    views at once & \multirow{2}{*}{anchor} & \multirow{2}{*}{key-rescaling} & \multirow{2}{*}{PSNR$\uparrow$} & \multirow{2}{*}{SSIM$\uparrow$} & \multirow{2}{*}{LPIPS$\downarrow$}\tabularnewline
    (100 at all) &  &  &  &  & \tabularnewline
    \hline 
    8/100 &  &  & 15.095 & 0.301 & 0.380\tabularnewline
    16/100 &  &  & 15.093 & 0.303 & 0.372\tabularnewline
    28/100 &  &  & 14.833 & 0.304 & 0.377\tabularnewline
    36/100 &  &  & 14.776 & 0.304 & 0.381\tabularnewline
    52/100 &  &  & 14.822 & 0.300 & 0.386\tabularnewline
    100/100 &  &  & 14.751 & 0.295 & 0.402\tabularnewline
    28/100 & $\checkmark$ &  & 13.879 & 0.253 & 0.469\tabularnewline
    100/100 &  & $\checkmark$ & \textbf{15.640} & \textbf{0.317} & \textbf{0.358}\tabularnewline
    \hline 
    \end{tabular}
    \vspace{-0.125in}
    \caption{\textbf{Ablation studies of expanding NVS to larger target view numbers.} Results are based on MipNeRF-360.\label{tab:abaltion_view_num}}
  \vspace{-0.2in}
\end{table}

\begin{figure}
\begin{center}
\includegraphics[width=0.95\linewidth]{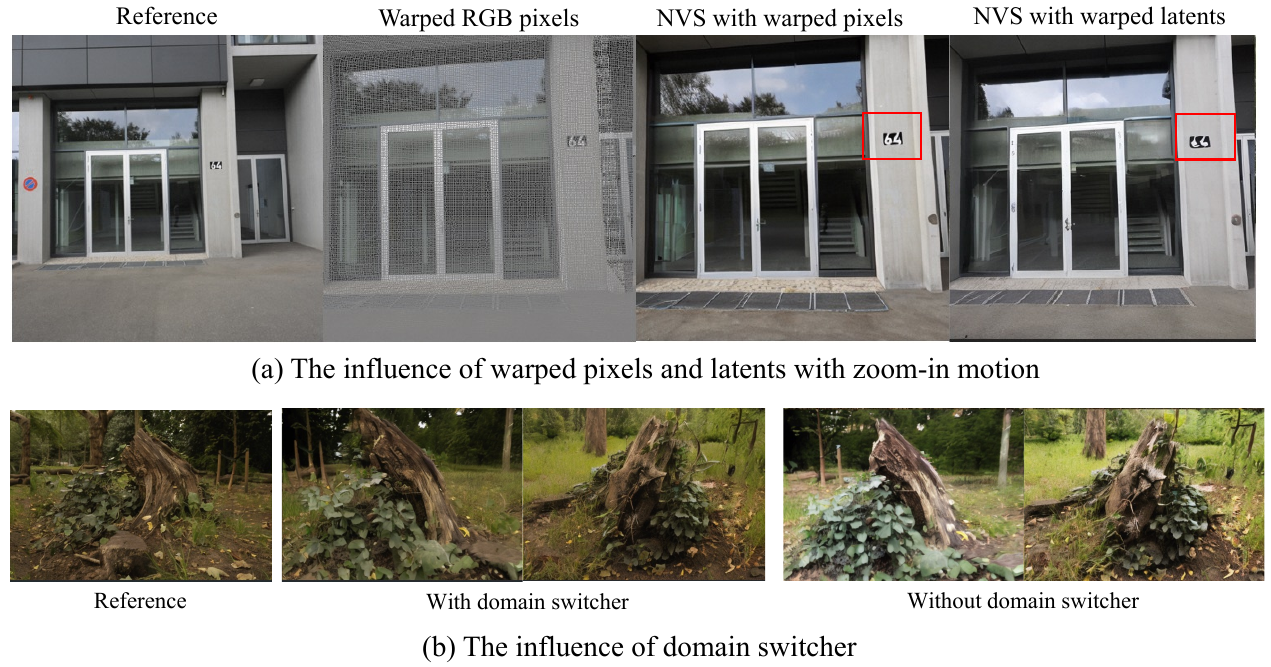}
\vspace{-0.25in}
\end{center}
   \caption{\textbf{Illustration of (a) warped pixels vs latents, (b) domain switcher.} The generated zoom-in number cards are framed in red.
   \label{fig:ablation_vis}}
\vspace{-0.3in}
\end{figure}

\subsection{Ablation Study}

Ablation studies on the training subset are listed in \Cref{tab:ablation1}, verifying the effectiveness of camera embedding, qk-norm, 3D priors, and the incorporation way of them.
Firstly, qk-norm enhances the performance with stable training convergence. CCM largely boosts the NVS results with global 3D positions. However, we observed that injecting CCM features through cross-attention modules struggles to achieve proper improvements compared to simply adding them to input features via convolutional encoding. 
This indicates that directly fusing pixel-wise aligned features is an effective way of learning 3D priors by self-attention.
Moreover, warped pixels and latents are comparable, but the former is more lightweight and achieves better details during the zoom-in camera motion as in \Cref{fig:ablation_vis}(a). 
Finally, camera embedding is still effective when 3D priors are incorporated into our model.
We further verify some training techniques' impact on the full training set in \Cref{tab:ablation_scaling_up}. Both multi-scale training and EMA strengthen NVS results.
Particularly, the domain switcher alleviates the color difference caused by inconsistent Megascenes images, as illustrated in \Cref{fig:ablation_vis}(b).

\noindent\textbf{View Extension.}
We evaluate the view extension ability in \Cref{tab:abaltion_view_num}. Three views serve as conditions, while others are generated at once. When view number $N+M>28$, the NVS results degraded significantly as discussed in \Cref{sec:key_rescale}. 
Additionally, we test the anchor-based NVS~\cite{gao2024cat3d}, \emph{i.e.}, generating 25 target views as anchors at first, followed by an iterative synthesis for surrounding views. However, this way achieved suboptimal results due to the accumulation of artifacts.
The proposed key-rescaling enables NVS with extensive and coherent target views without degradation. 

\noindent\textbf{Dataset Scalability.}
We explore the impact of the dataset scalability for MVGenMaster training in \Cref{tab:dataset_scalability}. Our analysis begins with a base model trained with the subset of the complete training set, including Co3Dv2, MVImgNet, DL3DV, GL3D, Scannet++, and 3D-Front.
Subsequently, the base model is fine-tuned using different additional datasets with the same steps. Note that we employ the domain switcher for Objaverse and Megascenes.
The results indicate that the training of NVS is data-hungry, while diverse datasets with various camera trajectories facilitate the scaling up of MVGenMaster on the zero-shot benchmark.

\section{Conclusion}

In this paper, we present MVGenMaster as a general framework to handle various NVS downstream tasks. 
Specifically, we employ the 3D priors achieved by metric depth and geometric warping to substantially strengthen NVS performance.
Moreover, MVGenMaster enjoys a flexible pipeline to address the synthesis of variable reference and target views enhanced by the key-rescaling technique.
To support the scalability of our model, we have developed a large multi-view dataset called MvD-1M, complemented by effective training strategies aimed at improving NVS outcomes.

% \noindent\textbf{Limitation.} Although MVGenMaster is a powerful NVS model, an interesting future work involves modifying the base model design, \emph{e.g.}, replacing the backbone with Diffusion Transformer (DiT)~\cite{peebles2023scalable} for superior capacity.
% Besides, unifying both rendering and multi-view generation is a promising way to improve consistency and stability of NVS.

{
    \small
    \bibliographystyle{ieeenat_fullname}
    \bibliography{main}
}

\newpage
\appendix

\section{Dataset Details}

\begin{figure*}
\begin{center}
\includegraphics[width=1.0\linewidth]{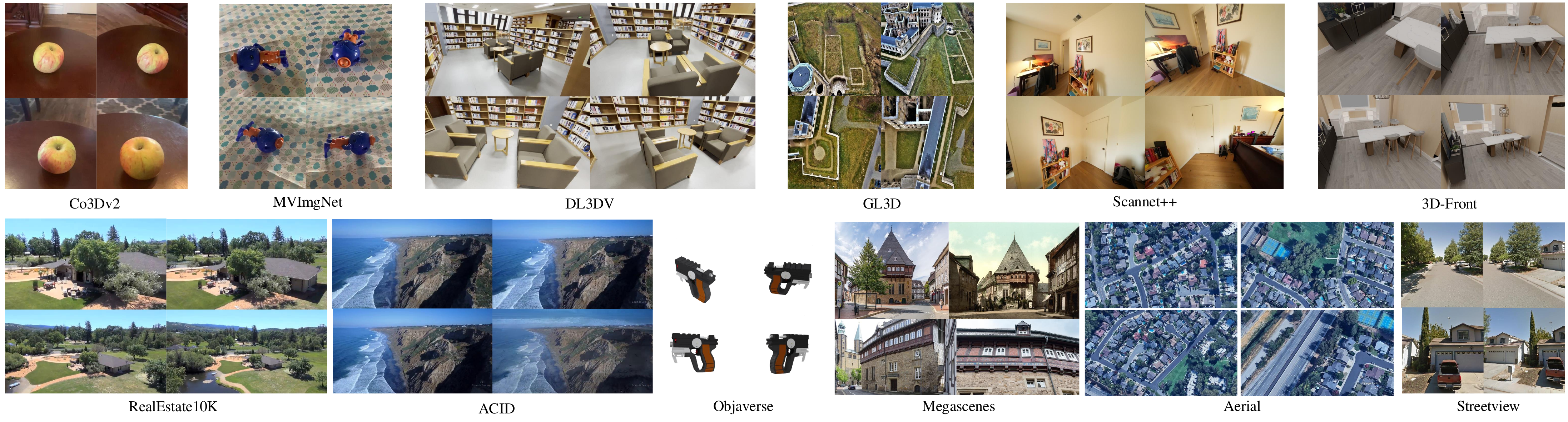}
\vspace{-0.3in}
\end{center}
   \caption{\textbf{Illustration of MvD-1M used in MVGenMaster.} 
   \label{fig:datasets}}
\vspace{-0.2in}
\end{figure*}

We discuss more details about the MvD-1M used in this work and the way to get metric depth of them. The visualization of some examples from MvD-1M is shown in \Cref{fig:datasets}.

\noindent\textbf{Co3Dv2~\cite{reizenstein21co3d}:} Co3Dv2 is a widely used object-centric dataset with diverse classes, backgrounds, and image resolutions. 
We filtered the official Co3Dv2 dataset, retaining the scenes whose short sides are larger than 256. 
Furthermore, we ensure that all categories in Co3Dv2 are sampled in balance during training.
Since Co3Dv2 contained sparse depth maps from SfM, which were used to align metric depth.

\noindent\textbf{MVImgNet~\cite{yu2023mvimgnet}:} MVImgNet contains high-quality object-centric images, while the viewpoint changes are not as large as Co3Dv2. Similarly, we maintained category balance for each training epoch in MVImgNet.
MVImgNet has sparse Colmap points which could be utilized to align metric depth.

\noindent\textbf{DL3DV~\cite{ling2024dl3dv}:} DL3DV is a large-scale, scene-level dataset with a variety of scenarios, and it serves as the primary source for our model's training. Note that only some DL3DV scenes contain sparse Colmap points, others do not. 
For scenes lacking SfM results, we employed the MVS~\cite{cao2024mvsformer++} to achieve metric depths, applying a confidence filter of $>0.5$.

\noindent\textbf{GL3D~\cite{shen2018mirror}:} GL3D primarily contains aerial images, alongside a limited number of scene-level multi-view images. Given that the authors did not provide detailed camera poses for the undistorted images, we utilized Colmap to estimate the camera poses and extract sparse SfM points.

\noindent\textbf{Scannet++~\cite{yeshwanthliu2023scannetpp}:} Scannet++ comprises indoor scenes. 
To ensure the image quality, we use images captured by DSLRs rather than RGBD from cell phones. 
We first undistorted the DSLR images, and then applied the officially provided sparse Colmap points to align the metric depth.

\noindent\textbf{3D-Front~\cite{fu20213d}:} 3D-Front includes numerous indoor scenes rendered using Blender. For this dataset, we directly utilized the rendered ground truth depth for geometric warping.

\noindent\textbf{RealEstate10K~\cite{zhou2018Stereo}:} RealEstate10K is a widely used scene-level dataset. We combined the training and test sets to maximize its utility, retaining 50 scenes for validation, as detailed in Table 1 of the main paper. Because RealEstate10k only contains camera poses, we employed MVS~\cite{cao2024mvsformer++} to get metric depths.

\noindent\textbf{ACID~\cite{liu2021infinite}:} ACID is built with aerial images captured in natural landscapes. We filtered out samples with minimal motion and used MVS~\cite{cao2024mvsformer++} to obtain the metric depths.

\noindent\textbf{Objaverse~\cite{deitke2023objaverse}:} All images from Objaverse are rendered in $512\times512$. To improve the diversity, we randomly altered the background colors for different objects in this dataset. Importantly, we did not apply 3D priors in this relatively simpler dataset to increase the challenge.

\noindent\textbf{Megascenes~\cite{tung2024megascenes}:} Megascenes contains diverse places collected from the wiki, captured at various times, using a range of devices.
We retained scenes with a valid view count of at least 8 and aligned the metric depth using sparse SfM points.

\noindent\textbf{Aerial:} The Aerial dataset was collected from Google Earth~\cite{googleearth}, capturing images below 500 meters above the ground. We retain the images with top-3 heights to ensure superior visual quality. To derive metric depths, we trained Instant-NGP~\cite{mueller2022instant} for each scene and rendered the metric depth through the reconstructed mesh.

\noindent\textbf{Streetview:} The Streetview dataset was also collected from Google Earth~\cite{googleearth}, which comprises both indoor and outdoor scenes from New York. All Streetview scenes are panoramic images. We utilized Equirectangular projection to derive standard images and camera poses with random intrinsic and extrinsic matrices to enhance diversity. 
Notably, as depth data is essentially meaningless for geometric warping for panoramic images, we opted not to employ geometric warping for this dataset, including only the camera poses.

\section{More Implementation Details}

\subsection{Canonical Coordinate Map (CCM)}

We provide more details about CCM~\cite{li2023sweetdreamer} mentioned in the main paper.
To incorporate precise positional information to MVGenMaster, we begin by unprojecting 2D coordinates $x_r$ from reference views into the 3D world space (canonical space) as:
 \begin{equation}
\label{eq:ccm}
C^{world}_r\simeq P^{c \rightarrow w}_r\hat{D}_r(x_r)K^{-1}_r x_r,
\end{equation}
where $C^{world}_r$ denotes the 3D coordinates in the world space (CCM); $P^{c \rightarrow w}_r$ is the transformation matrix converting coordinates from camera to world space; $\hat{D}_r(x_r)$ and $K^{-1}_r$ indicate metric depth and camera intrinsic's inversion of reference views, respectively.
Subsequently, $C^{world}_r$ is further warped into target views, \ie, $C^{warp}_{r \rightarrow t}$ following Eq.~\ref{eq:depth_align} in the main paper.
Note that we further normalize $C^{warp}_{r \rightarrow t}$ into 0 to 1 before send it to the Fourier embedding.

\subsection{Multi-Scale Training}

All resolutions used in our multi-scale training are listed in \Cref{tab:multi_scale}. And we ensure that at least one batch samples are allocated for each resolution group per-epoch.

\begin{table}[h!]
  \small
  \centering
\begin{tabular}{ccc}
\hline 
\multicolumn{3}{c}{Resolution (height $\times$ width)}\tabularnewline
\hline 
320$\times$768 & 384$\times$448 & 576$\times$448\tabularnewline
320$\times$704 & 448$\times$576 & 576$\times$384\tabularnewline
320$\times$640 & 448$\times$512 & 576$\times$320\tabularnewline
320$\times$576 & 512$\times$512 & 640$\times$384\tabularnewline
320$\times$512 & 448$\times$384 & 640$\times$320\tabularnewline
384$\times$640 & 512$\times$448 & 704$\times$320\tabularnewline
384$\times$576 & 512$\times$384 & 768$\times$320\tabularnewline
384$\times$512 & 512$\times$320 & \tabularnewline
\hline 
\end{tabular}
    \vspace{-0.1in}
    \caption{\textbf{Resolutions used in the multi-scale training.\label{tab:multi_scale}}}
  \vspace{-0.1in}
\end{table}

\subsection{Normalization of Camera Poses}

The Pl{\"u}cker ray~\cite{xu2023dmv3d} enjoys dense and good camera presentations for NVS, which can be denoted as $(\mathbf{o}\times\mathbf{d},\mathbf{d})$, where $\mathbf{o},\mathbf{d}$ indicate the origin and direction of pixel-aligned rays respectively. 
We clarify that the cross-product result $\mathbf{o}\times\mathbf{d}$ is not scaling invariable due to different scales of $\mathbf{o}$.
Since the SfM results of different datasets contain different camera scales, we need to normalize them before the training and inference to avoid unseen camera scales during the inference. Specifically, we analyzed all camera positions within each scene and saved the longest distance between the two farthest cameras. If this distance exceeded 5, we normalized the global scene to ensure the longest side was 5, while the metric depth should also be re-scaled accordingly.
Such normalization allows the model to effectively adapt small camera motions instead of straightforwardly normalizing all cameras into the same scale.

\subsection{Settings of 3DGS Reconstruction}

We follow the vanilla 3DGS implementation of~\cite{kerbl20233d} with L1, SSIM, and LPIPS~\cite{zhang2018unreasonable} losses as many other works~\cite{melas20243d,liu2024reconx,yu2024viewcrafter}. The point cloud from Dust3R~\cite{wang2024dust3r} is leveraged to initialize the 3DGS training. The dynamic LPIPS weighting strategy~\cite{liu2024reconx,gao2024cat3d} is also incorporated according to the distance to the nearest reference view, and the LPIPS weight changes from 0 (reference view) to 0.25 (the farthest view). The training is accomplished within 2000 steps in our work.

\subsection{Efficiency and Memory Costs}

Our model enjoys a similar efficiency as CAT3D~\cite{gao2024cat3d}, while the cost of 3D priors incorporation is negligible. 
Inference efficiency and GPU memory costs of the 50-step DDIM scheduler with different target view numbers are listed in \Cref{tab:efficiency}. The experimental device is one 80GB A800 NVIDIA GPU; the image resolution is 384$\times$576. Note that our model can synthesize all views at once instead of cumbersome iterative generations.

\begin{table}[h]
\small
\centering
\begin{tabular}{cccc}
\hline 
Views & 3D priors & Time & Memory\tabularnewline
\hline 
8 & $\times$ & 7.0085s & 6420M\tabularnewline
8 & $\checkmark$ & 7.1768s & 6440M\tabularnewline
25 & $\checkmark$ & 21.272s & 14048M\tabularnewline
50 & $\checkmark$ & 53.055s & 27762M\tabularnewline
100 & $\checkmark$ & 132.67s & 50542M\tabularnewline
158 & $\checkmark$ & 262.86s & 80738M\tabularnewline
\hline 
\end{tabular}
\vspace{-0.1in}
\caption{\textbf{Efficiency and memory costs of our model.} The experimental device is one 80GB NVIDIA A800 GPU, while the image resolution is 384x576. The setting of 8-view without 3D priors can be regarded as the baseline CAT3D~\cite{gao2024cat3d}.\label{tab:efficiency}}
\vspace{-0.1in}
\end{table}

\section{Supplementary Experiment Results}

\begin{figure}[h!]
\begin{center}
\includegraphics[width=1.0\linewidth]{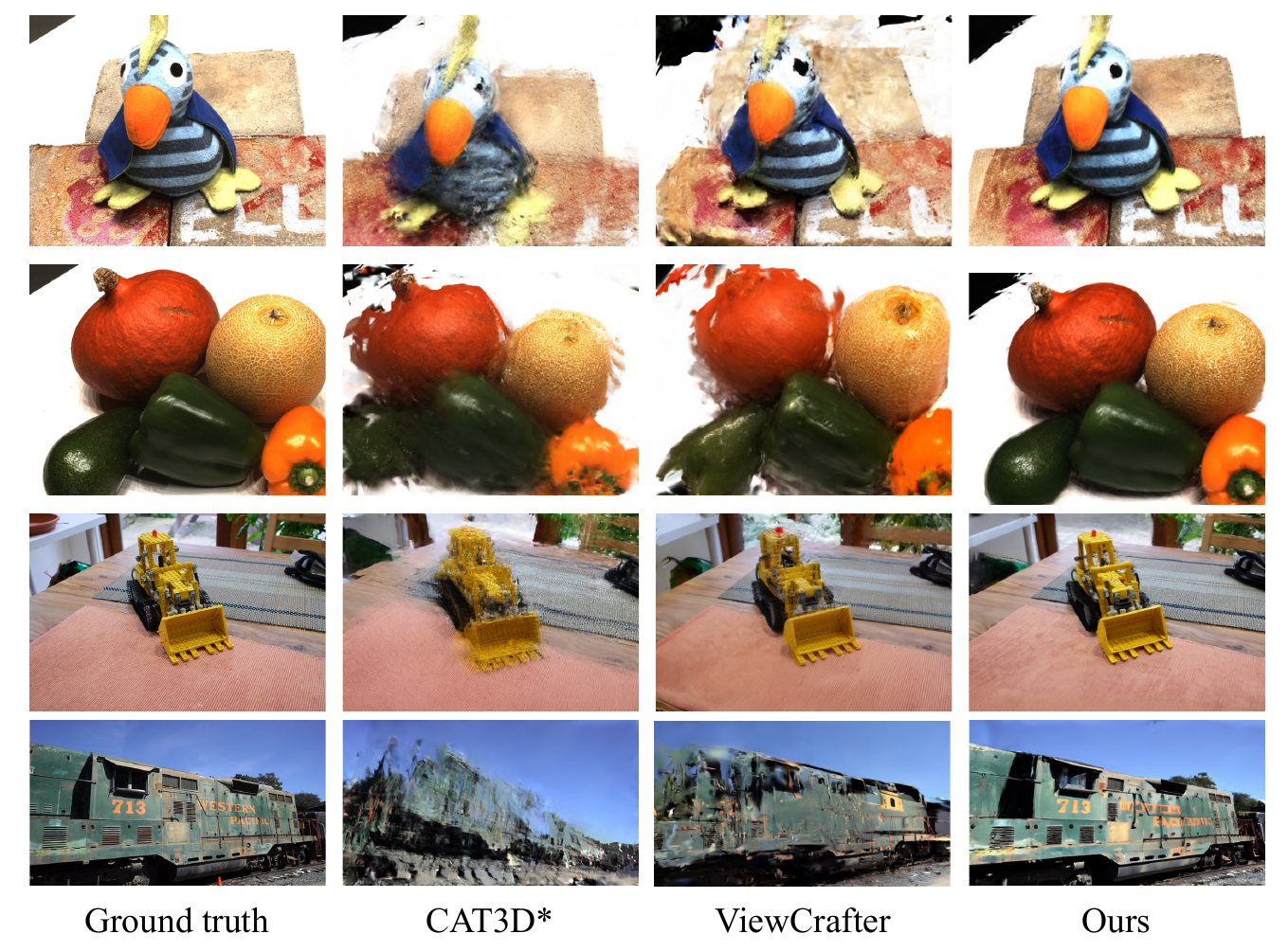}
\vspace{-0.3in}
\end{center}
   \caption{\textbf{Visualization of the novel views from 3DGS reconstruction on DTU, MipNeRF-360, and Tanks-and-Temples.} 3DGS results are trained with 21 frames, while the other 4 frames are validated. Given the first and last frames, other views are generated by related methods.
   \label{fig:gs_result}}
\vspace{-0.15in}
\end{figure}

\begin{figure*}
\begin{center}
\includegraphics[width=1.0\linewidth]{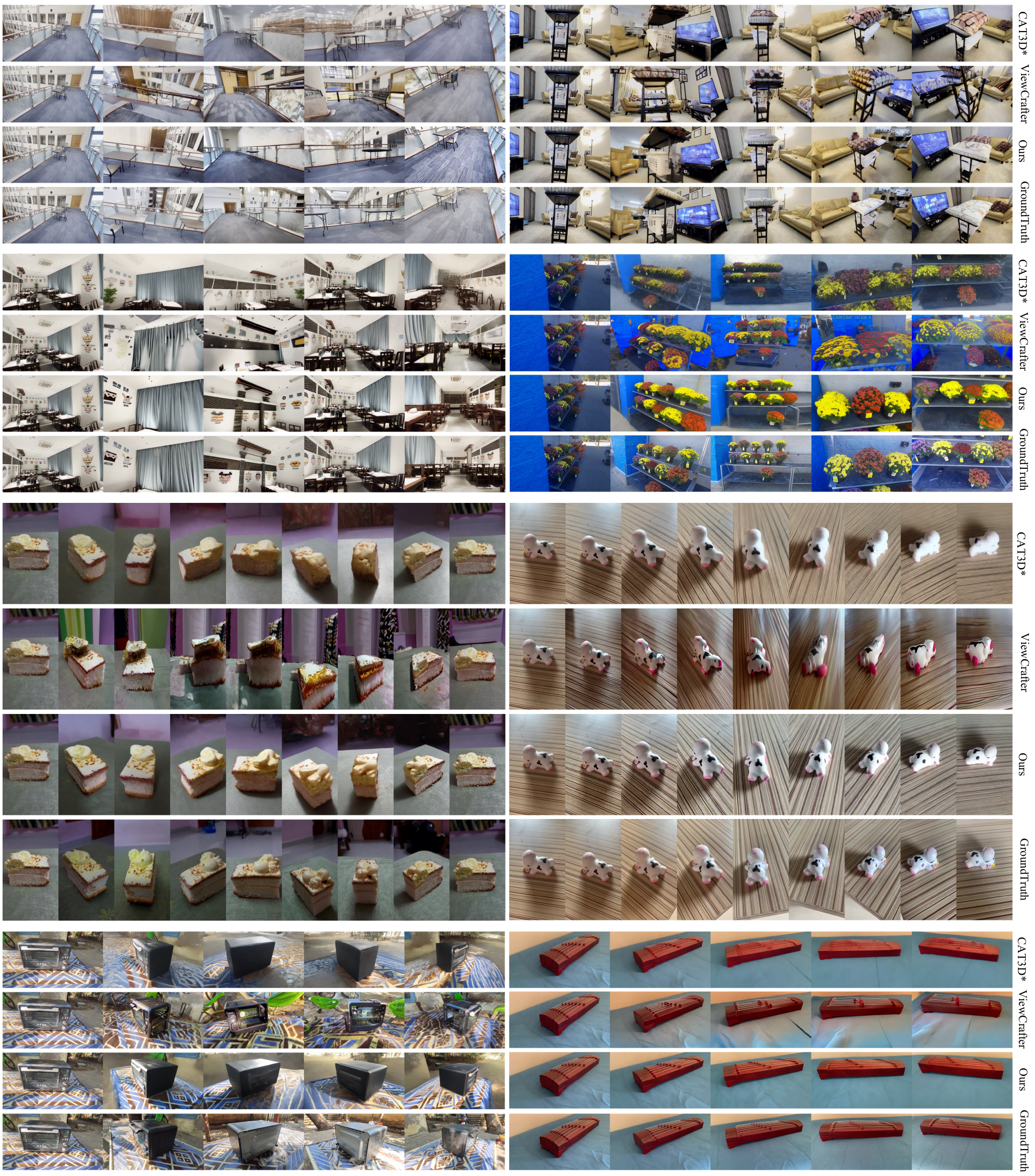}
\vspace{-0.25in}
\end{center}
   \caption{\textbf{Qualitative NVS results compared among CAT3D*, ViewCrafter, and our MVGenMaster from DL3DV, MVImgNet, and Co3Dv2.} The synthesis is based on ($N=1$) reference view and ($M=24$) target views.
   The leftmost column displays the reference view, while the remaining visualizations are uniformly sampled from the 24-frame generation due to page limitation. 
   \label{fig:nvs_qualitative2}}
\vspace{-0.15in}
\end{figure*}

\begin{figure*}
\begin{center}
\includegraphics[width=1.0\linewidth]{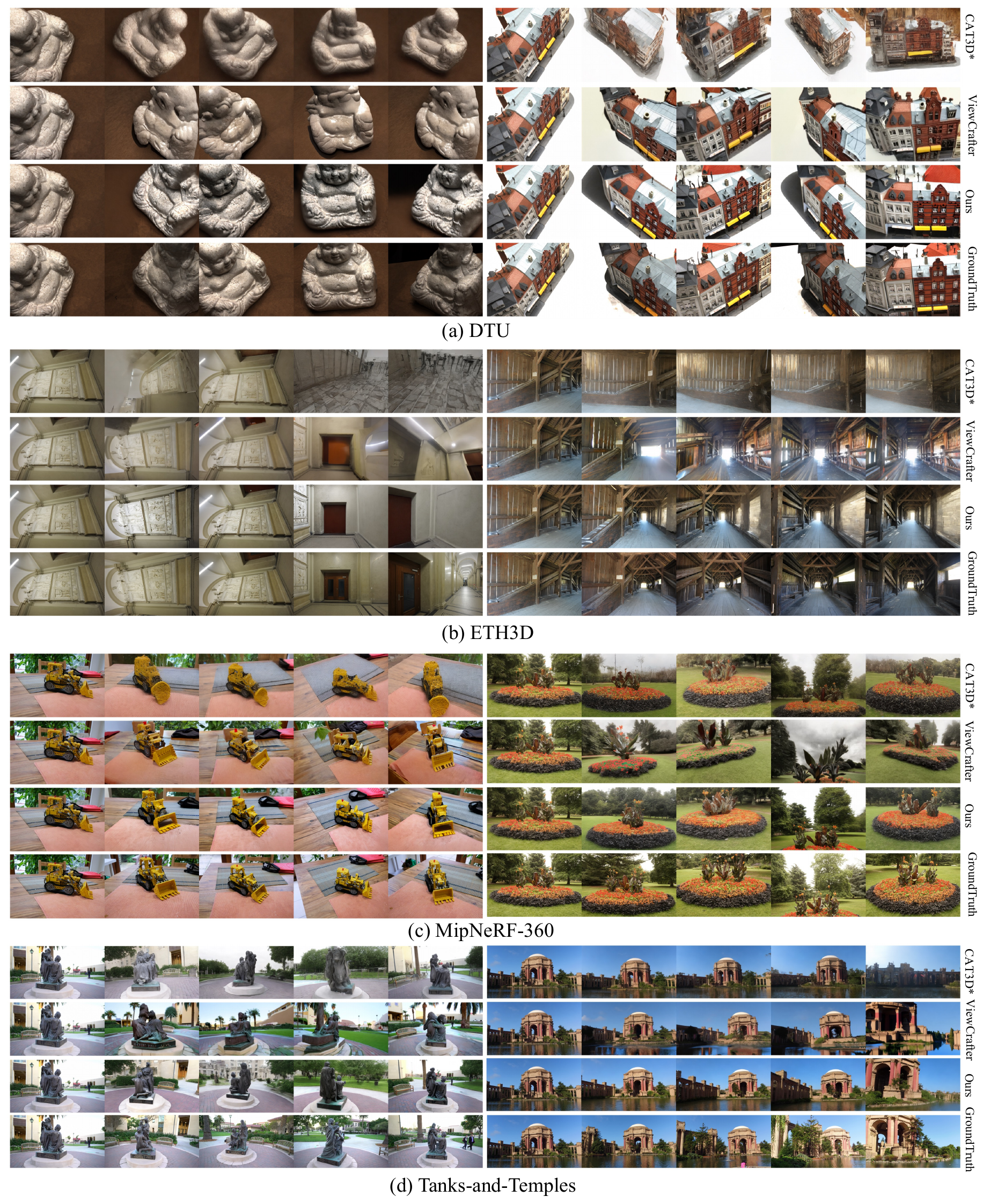}
\vspace{-0.35in}
\end{center}
   \caption{\textbf{Qualitative NVS results compared among CAT3D*, ViewCrafter, and our MVGenMaster from zero-shot datasets.} The synthesis is based on ($N=1$) reference view and ($M=24$) target views.
   The leftmost column displays the reference view, while the remaining visualizations are uniformly sampled from the 24-frame generation due to page limitation. 
   \label{fig:nvs_qualitative3}}
\vspace{-0.15in}
\end{figure*}

\begin{figure*}
\begin{center}
\includegraphics[width=1.0\linewidth]{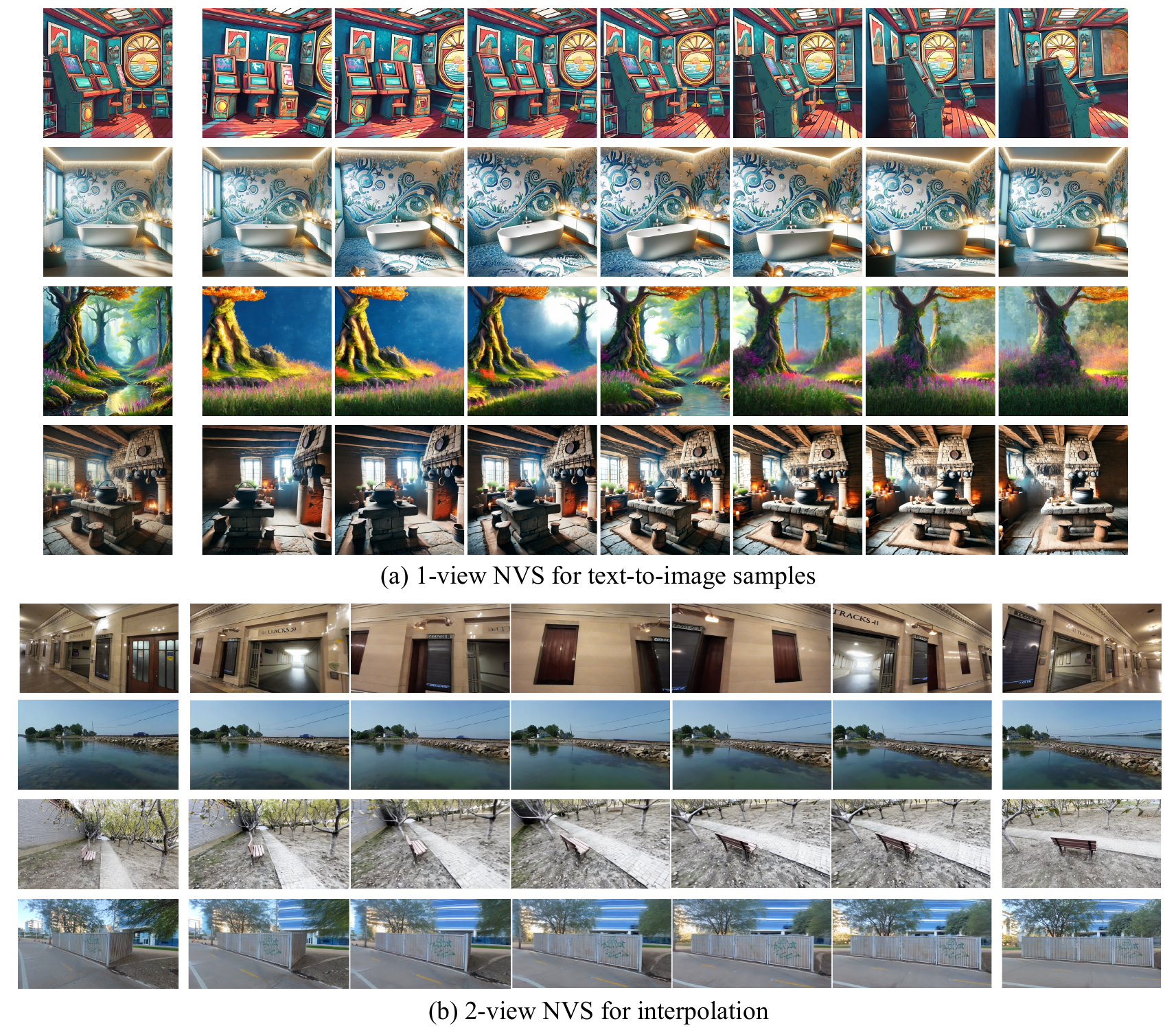}
\vspace{-0.35in}
\end{center}
   \caption{\textbf{More results of MVGenMaster based on 1-view NVS and 2-view interpolation.} 
   \label{fig:nvs_application2}}
\vspace{-0.15in}
\end{figure*}

\begin{figure*}
\begin{center}
\includegraphics[width=0.95\linewidth]{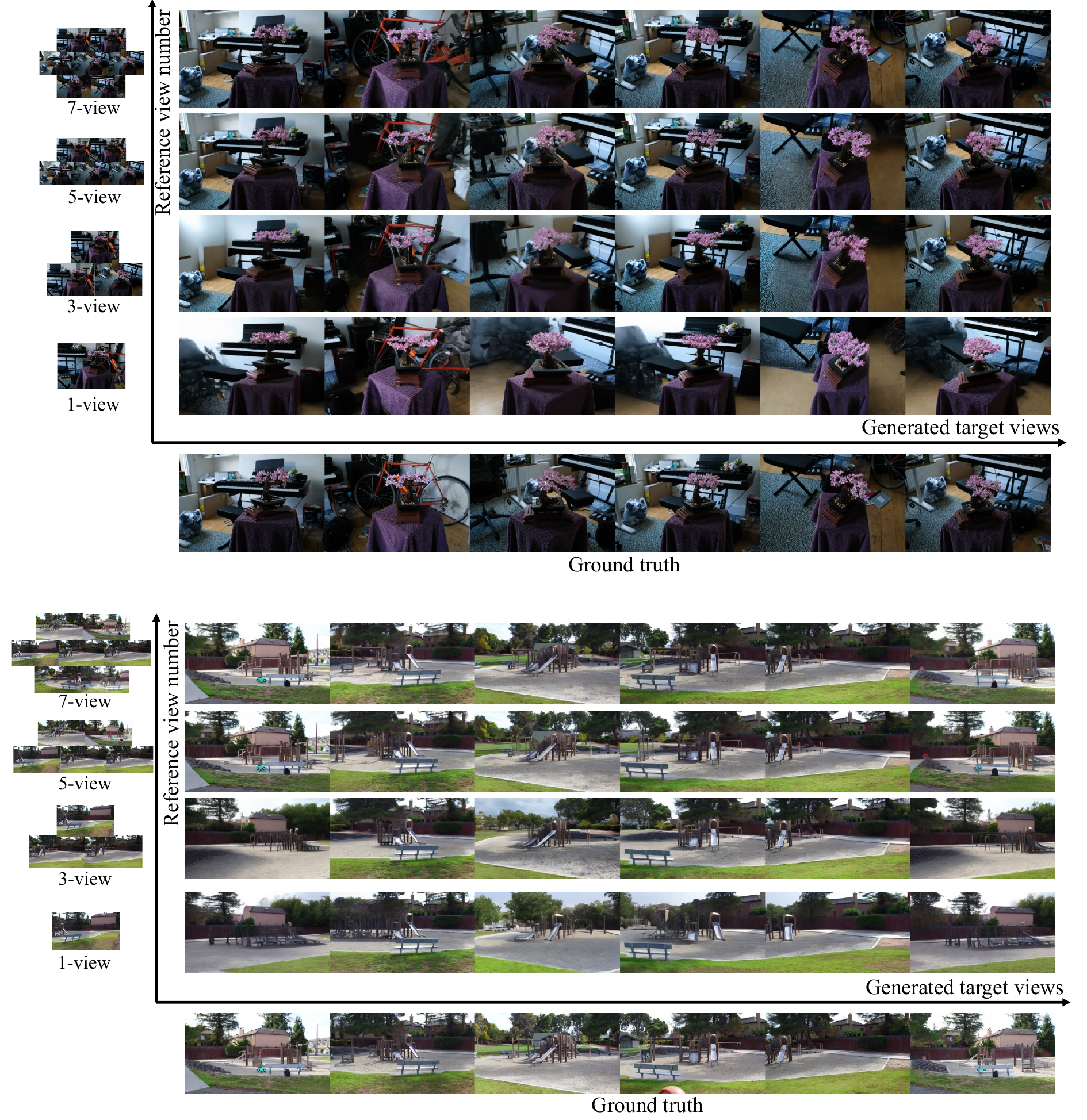}
\vspace{-0.25in}
\end{center}
   \caption{\textbf{NVS results with varying reference views.} 
   \label{fig:diff_ref_results}}
\vspace{-0.15in}
\end{figure*}

We show the qualitative comparison of 3DGS reconstruction in \Cref{fig:gs_result}. More NVS results are listed in \Cref{fig:nvs_qualitative2}, \Cref{fig:nvs_qualitative3}, and \Cref{fig:nvs_application2}. Additionally, more reference views further strengthen the performance as shown in \Cref{tab:diff_ref_view} and \Cref{fig:diff_ref_results}.

\begin{table}[h]
\footnotesize
\centering
{\footnotesize{}}%
\setlength{\tabcolsep}{3pt} % 减小列间距
\begin{tabular}{c|c|cc|cc|cc}
\hline 
 &  & \multicolumn{2}{c|}{{\footnotesize{}3-view}} & \multicolumn{2}{c|}{{\footnotesize{}6-view}} & \multicolumn{2}{c}{{\footnotesize{}9-view}}\tabularnewline
 &  & {\footnotesize{}PSNR$\uparrow$} & {\footnotesize{}LPIPS$\downarrow$} & {\footnotesize{}PSNR$\uparrow$} & {\footnotesize{}LPIPS$\downarrow$} & {\footnotesize{}PSNR$\uparrow$} & {\footnotesize{}LPIPS$\downarrow$}\tabularnewline
\hline 
\multirow{3}{*}{\begin{turn}{90}
{\footnotesize{}Real10k}
\end{turn}} & {\footnotesize{}Recon\cite{wu2024reconfusion}} & {\footnotesize{}\underline{25.84}} & {\footnotesize{}0.144} & {\footnotesize{}\underline{29.99}} & {\footnotesize{}0.103} & {\footnotesize{}\underline{31.82}} & {\footnotesize{}0.092}\tabularnewline
 & {\footnotesize{}CAT3D\cite{gao2024cat3d}} & \textbf{\footnotesize{}26.78} & {\footnotesize{}\underline{0.132}} & \textbf{\footnotesize{}31.07} & {\footnotesize{}\underline{0.092}} & \textbf{\footnotesize{}32.20} & {\footnotesize{}\underline{0.082}}\tabularnewline
 & {\footnotesize{}Ours} & {\footnotesize{}25.15} & \textbf{\footnotesize{}0.091} & {\footnotesize{}27.28} & \textbf{\footnotesize{}0.067} & {\footnotesize{}27.54} & \textbf{\footnotesize{}0.062}\tabularnewline
\hline 
\multirow{3}{*}{\begin{turn}{90}
{\footnotesize{}CO3D}
\end{turn}} & {\footnotesize{}Recon\cite{wu2024reconfusion}} & {\footnotesize{}19.59} & {\footnotesize{}0.398} & {\footnotesize{}21.84} & {\footnotesize{}0.342} & {\footnotesize{}\underline{22.95}} & {\footnotesize{}0.318}\tabularnewline
 & {\footnotesize{}CAT3D\cite{gao2024cat3d}} & \textbf{\footnotesize{}20.57} & {\footnotesize{}\underline{0.351}} & \textbf{\footnotesize{}22.79} & {\footnotesize{}\underline{0.292}} & \textbf{\footnotesize{}23.58} & {\footnotesize{}\underline{0.273}}\tabularnewline
 & {\footnotesize{}Ours} & {\footnotesize{}\underline{19.99}} & \textbf{\footnotesize{}0.278} & {\footnotesize{}\underline{21.96}} & \textbf{\footnotesize{}0.214} & {\footnotesize{}22.77} & \textbf{\footnotesize{}0.195}\tabularnewline
\hline 
\multirow{3}{*}{\begin{turn}{90}
{\footnotesize{}Mip360}
\end{turn}} & {\footnotesize{}Recon\cite{wu2024reconfusion}} & {\footnotesize{}15.50} & {\footnotesize{}0.585} & {\footnotesize{}16.93} & {\footnotesize{}0.544} & {\footnotesize{}18.19} & {\footnotesize{}0.511}\tabularnewline
 & {\footnotesize{}CAT3D\cite{gao2024cat3d}} & \textbf{\footnotesize{}16.62} & {\footnotesize{}\underline{0.515}} & \textbf{\footnotesize{}17.72} & {\footnotesize{}\underline{0.482}} & \textbf{\footnotesize{}18.67} & {\footnotesize{}\underline{0.460}}\tabularnewline
 & {\footnotesize{}Ours} & {\footnotesize{}\underline{16.29}} & \textbf{\footnotesize{}0.489} & {\footnotesize{}\underline{17.51}} & \textbf{\footnotesize{}0.405} & {\footnotesize{}\underline{18.21}} & \textbf{\footnotesize{}0.354}\tabularnewline
\hline 
\end{tabular}{\footnotesize\par}
\vspace{-0.1in}
\caption{\textbf{Reconstruction results of the official Reconfusion benchmark.} Results of reconfusion~\cite{wu2024reconfusion} and CAT3D~\cite{gao2024cat3d} are from their papers. Our results are under once NVS inference and 3DGS.\label{tab:reconfusion}}
\vspace{-0.15in}
\end{table}

\noindent\textbf{Reconstruction Benchmark.}
To ensure the effectiveness of MVGenMaster, we compare our method with the official data splits from the benchmark of Reconfusion~\cite{wu2024reconfusion} as in \Cref{tab:reconfusion}. Results of Reconfusion~\cite{wu2024reconfusion} and CAT3D~\cite{gao2024cat3d} are from their papers. Our method enjoys a much simpler pipeline to produce all views with one forward process without iterative generation (CAT3D) and distillation (Reconfusion), only taking 3 minutes for each scene with one A800 (NVS+3DGS). Besides, Reconfusion and CAT3D use ZipNeRF~\cite{barron2023zip}, which costs much more training times for each instance.
Moreover, we should clarify that the results of MipNeRF-360 are slightly different from Table.3 of the main paper. Because the data splits of Reconfusion are different from ours. They use a heuristic strategy to encourage more reasonable camera placing for the object centric reconstruction~\cite{wu2024reconfusion}.
Even under such a challenging setting, our method achieves better human perception (lower LPIPS) and comparable PSNR.

\begin{table}[h]
  \small
  \centering
    \begin{tabular}{l|ccc}
    \hline 
    \textbf{Datasets} & \multirow{2}{*}{PSNR$\uparrow$} & \multirow{2}{*}{SSIM$\uparrow$} & \multirow{2}{*}{LPIPS$\downarrow$}\tabularnewline
    Reference View &  &  & \tabularnewline
    \hline 
    \textbf{CO3D+MVImgNet} &  &  & \tabularnewline
    Ours (1-view) & 18.619 & 0.573 & 0.316\tabularnewline
    Ours (3-view) & 21.466 & 0.653 & 0.220\tabularnewline
    Ours (5-view) & 22.594 & 0.682 & 0.188\tabularnewline
    Ours (7-view) & \textbf{23.299} & \textbf{0.697} & \textbf{0.169}\tabularnewline
    \hline 
    \textbf{DL3DV+Real10k} &  &  & \tabularnewline
    Ours (1-view) & 15.729 & 0.468 & 0.376\tabularnewline
    Ours (3-view) & 18.296 & 0.552 & 0.266\tabularnewline
    Ours (5-view) & 19.366 & 0.585 & 0.224\tabularnewline
    Ours (7-view) & \textbf{20.039} & \textbf{0.604} & \textbf{0.204}\tabularnewline
    \hline 
    \textbf{Zero-shot Datasets} &  &  & \tabularnewline
    Ours (1-view) & 12.879 & 0.394 & 0.466\tabularnewline
    Ours (3-view) & 15.533 & 0.491 & 0.319\tabularnewline
    Ours (5-view) & 16.826 & 0.531 & 0.263\tabularnewline
    Ours (7-view) & \textbf{17.559} & \textbf{0.550} & \textbf{0.233}\tabularnewline
    \hline 
    \end{tabular}
  \vspace{-0.1in}
  \caption{\textbf{Results of MVGenMaster with different reference views.} The total view number $N+M=25$. \label{tab:diff_ref_view}}
  \vspace{-0.15in}
\end{table}

\begin{table}[h!]
\small
\centering
\begin{tabular}{c|c|ccc}
\hline 
\multirow{2}{*}{views at once} & \multirow{2}{*}{key-rescaling} & \multirow{2}{*}{PSNR$\uparrow$} & \multirow{2}{*}{SSIM$\uparrow$} & \multirow{2}{*}{LPIPS$\downarrow$}\tabularnewline
 &  &  &  & \tabularnewline
\hline 
8/158 & -- & 15.193 & 0.378 & 0.362\tabularnewline
28/158 & -- & 14.913 & 0.386 & 0.361\tabularnewline
158/158 & 1.15 & 15.439 & \textbf{0.410} & 0.358\tabularnewline
158/158 & 1.2 & 15.565 & 0.409 & \textbf{0.355}\tabularnewline
158/158 & 1.25 & \textbf{15.597} & 0.403 & 0.356\tabularnewline
158/158 & 1.3 & 15.538 & 0.397 & 0.359\tabularnewline
\hline 
\end{tabular}
\vspace{-0.1in}
\caption{\textbf{Results of extremely long sequence generation of MVGenMaster.} The total view number $N+M=158$ reaches the upper bound of an 80GB GPU. \label{tab:ex_longseq}}
\vspace{-0.15in}
\end{table}

\noindent\textbf{Extending to More Target Views.}
To explore the robustness of the proposed key-rescaling, we conduct an exploratory ablation study in \Cref{tab:ex_longseq}. In this study, we increase the total number of views to 158, which represents the memory limit of an 80GB GPU. Key-rescaling with $\gamma=1.2$ is still effective and robust in persisting the reference guidance for such an extremely long sequence.
We should clarify that this memory limit is merely a hardware constraint of our method, which could potentially be overcome with efficient engineering optimizations. We regard these engineering optimizations as interesting future work to further extend the target view numbers of MVGenMaster. Furthermore, our method's capability to generate 100 views is sufficient to address most NVS scenarios, as verified in our main paper.

\begin{table}[h!]
  \small
  \centering
{\small{}}%
\setlength{\tabcolsep}{4pt} % 减小列间距
\begin{tabular}{c|ccc|c|ccc}
\hline 
{\small{}N/M} & {\small{}$\text{\ensuremath{\gamma}}$} & {\small{}PSNR$\uparrow$} & {\small{}LPIPS$\downarrow$} & {\small{}N/M} & {\small{}$\text{\ensuremath{\gamma}}$} & {\small{}PSNR$\uparrow$} & {\small{}LPIPS$\downarrow$}\tabularnewline
\hline 
\multirow{4}{*}{{\small{}1/97}} & {\small{}1.0} & {\small{}\underline{12.53}} & {\small{}0.541} & \multirow{4}{*}{{\small{}2/97}} & {\small{}1.0} & {\small{}13.69} & {\small{}0.468}\tabularnewline
 & {\small{}1.1} & \textbf{\small{}12.56} & {\small{}0.528} &  & {\small{}1.1} & {\small{}14.31} & {\small{}\underline{0.432}}\tabularnewline
 & {\small{}1.2} & {\small{}12.52} & \textbf{\small{}0.523} &  & {\small{}1.2} & \textbf{\small{}14.46} & \textbf{\small{}0.426}\tabularnewline
 & {\small{}1.3} & {\small{}12.50} & {\small{}\underline{0.526}} &  & {\small{}1.3} & {\small{}\underline{14.35}} & {\small{}0.438}\tabularnewline
\hline 
\multirow{4}{*}{{\small{}3/50}} & {\small{}1.0} & {\small{}14.82} & {\small{}0.386} & \multirow{4}{*}{{\small{}3/97}} & {\small{}1.0} & {\small{}14.75} & {\small{}0.402}\tabularnewline
 & {\small{}1.1} & {\small{}\underline{15.47}} & {\small{}\underline{0.359}} &  & \textbf{\small{}A1} & {\small{}\underline{15.59}} & {\small{}\underline{0.360}}\tabularnewline
 & {\small{}1.2} & \textbf{\small{}15.55} & \textbf{\small{}0.358} &  & {\small{}1.2} & \textbf{\small{}15.64} & \textbf{\small{}0.358}\tabularnewline
 & {\small{}1.3} & {\small{}15.28} & {\small{}0.371} &  & \textbf{\small{}A2} & {\small{}15.48} & {\small{}0.364}\tabularnewline
\hline 
\end{tabular}{\small\par}
  \vspace{-0.1in}
  \caption{\textbf{Detailed Ablation of Key-rescaling.} Following the adaptive $\gamma$ settings based on attention-entropy~\cite{jin2023training}, \textbf{A1} is set based on the length of $N+M$, while \textbf{A2} further considers the various sequential lengths across different layers.\label{tab:gamma_aba}}
  \vspace{-0.15in}
\end{table}

\noindent\textbf{Detailed Ablation of Key-rescaling.}
We clarified that $\gamma=1.2$ defined in the main paper works robustly enough for our method across various $N,M$ (up to upper bound $N+M$=158). We further provide detailed ablation studies and discussions about key-rescaling in \Cref{tab:gamma_aba}.
i) $N=1$ suffers from ambiguous NVS by generating vast number of views (usually set $M\leq25$ for $N=1$). As exploratory results, key-rescaling's effect of $N=1$ is a little weaker than $N>1$, but it still achieves the best quality under $\gamma=1.2$ (best LPIPS).
ii) Longer sequence enjoys larger $\gamma$, while shorter one requires smaller $\gamma$. Since $N+M=100$ addresses most scenarios, $\gamma=1.2$ is robust enough for our method.
iii) We show adaptive $\gamma$ settings based on attention-entropy~\cite{jin2023training} (\textbf{A1}, \textbf{A2}). They fail to get superior results compared to constant $\gamma$. 
Further exploring the key-rescaling theory and finding the optimal $\gamma$ is interesting future work. Our work provides this insight to the community and verifies the proposed key-rescaling is generalized enough to address our problems based on Occam's Razor. 

\begin{table}[h!]
\small
\centering
\begin{tabular}{c|ccc}
\hline 
\multirow{2}{*}{Methods} & \multirow{2}{*}{PSNR$\uparrow$} & \multirow{2}{*}{SSIM$\uparrow$} & \multirow{2}{*}{LPIPS$\downarrow$}\tabularnewline
 &  &  & \tabularnewline
\hline 
Warp+Pose~\cite{tung2024megascenes} (1-view) & 9.808 & 0.201 & 0.597\tabularnewline
MVGenMaster (1-view) & 11.188 & 0.299 & 0.575\tabularnewline
MVGenMaster (3-view) & \textbf{12.374} & \textbf{0.347} & \textbf{0.484}\tabularnewline
\hline 
\end{tabular}
\vspace{-0.1in}
\caption{\textbf{Results on the challenging Megascenes dataset.} Our method is compared with the warp+pose based model in~\cite{tung2024megascenes}.\label{tab:megascenes_exp}}
\vspace{-0.15in}
\end{table}

\begin{figure}
\begin{center}
\includegraphics[width=1.0\linewidth]{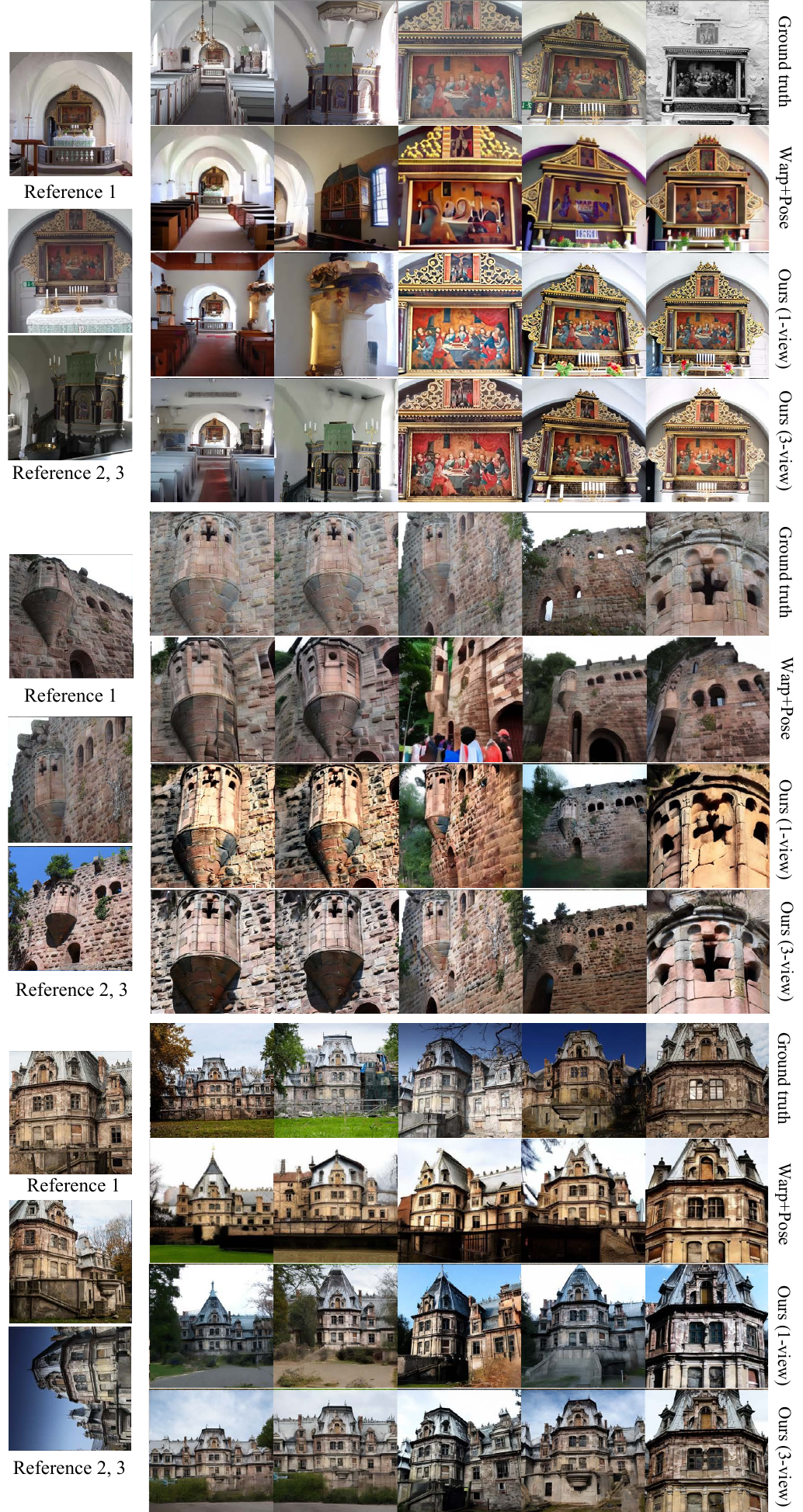}
\vspace{-0.3in}
\end{center}
   \caption{\textbf{Qualitative NVS results compared with the Warp+Pose baseline in~\cite{tung2024megascenes} on Megascenes.} 
   \label{fig:megascenes_nvs}}
\vspace{-0.15in}
\end{figure}

\noindent\textbf{Results Compared on Megascenes.}
We further compare the NVS results on the challenging Megascenes dataset~\cite{tung2024megascenes} in \Cref{tab:megascenes_exp} and \Cref{fig:megascenes_nvs}. Our method outperforms the Warp+Pose baseline in~\cite{tung2024megascenes} with 1-view reference. Moreover, our 3-view-based results achieve better qualities.
Although both methods of MVGenMaster and~\cite{tung2024megascenes} used 3D priors from depth estimation, MVGenMaster could handle high-resolution, multi-view consistent, and impressive generations with varying reference and target views. In contrast, the baseline in~\cite{tung2024megascenes} only performs with restricted low-resolution (256x256) and 1-view conditions.

\begin{figure*}
\begin{center}
\includegraphics[width=0.77\linewidth]{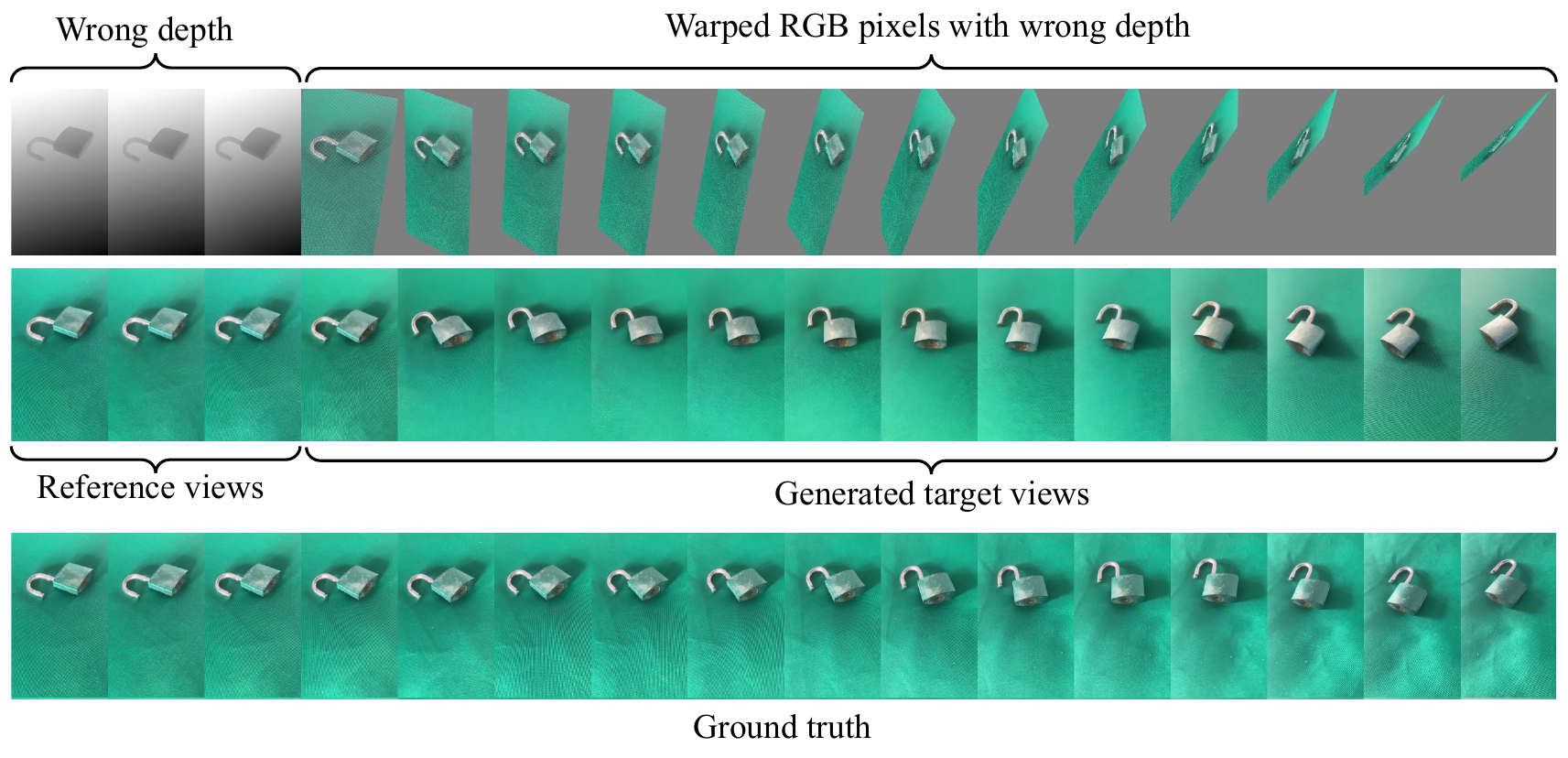}
\vspace{-0.25in}
\end{center}
   \caption{\textbf{Visualization of NVS results with inaccurate metric depth.} Our method still shows good robustness.
   \label{fig:rubost_3d_priors}}
\vspace{-0.15in}
\end{figure*}

\noindent\textbf{Robustness of 3D Priors.}
Benefiting from the 3D prior dropout and suboptimal monocular depth alignment, MVGenMaster enjoys good robustness to 3D priors as illustrated in \Cref{fig:rubost_3d_priors}. In this case, the depth alignment is failed, leading to poor metric depth maps. However, our method still achieves proper results with camera pose conditions.

\begin{figure}[h!]
\begin{center}
\includegraphics[width=1.0\linewidth]{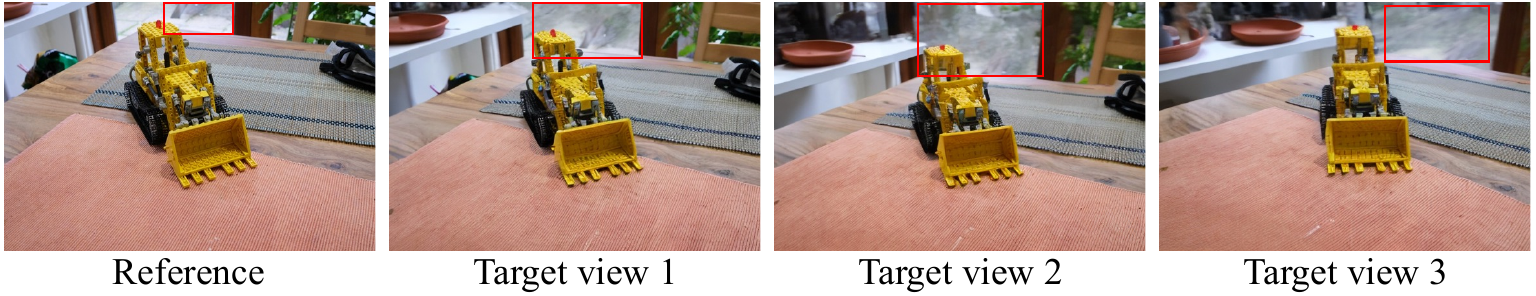}
\vspace{-0.3in}
\end{center}
   \caption{\textbf{Limitation.} The artifacts in the background caused by unclear and ambiguous background regions in the reference view. 
   \label{fig:limitation}}
\vspace{-0.15in}
\end{figure}

\noindent\textbf{Limitations.}
In \Cref{fig:limitation}, we show some ambiguous artifacts in the background generated by our model. Such artifacts are mainly caused by the unclear and ambiguous background regions of the reference view.
Improving the model backbone with superior capacity would refine this issue as mentioned in our main paper.

\section{Discussion and Future Work}
 Although MVGenMaster is a powerful NVS model, an interesting future work involves modifying the base model design, \emph{e.g.}, replacing the backbone with Diffusion Transformer (DiT)~\cite{peebles2023scalable} for superior capacity.
 Besides, unifying both rendering and multi-view generation is a promising way to improve consistency and stability of NVS.

\section{Broader Impacts}

This paper delves into the realm of image-based multi-view generation. Because of the powerful generative capacity, these models pose risks such as the potential for misinformation and the creation of fake images. 
We sincerely remind users to pay attention to generated content.
Besides, it is crucial to prioritize privacy and consent, as generative models frequently rely on vast datasets that may include sensitive information. Users must remain vigilant about these considerations to uphold ethical standards in their applications.
% Furthermore, generative models may perpetuate some biases according to the training data, leading to unfair outcomes. 
% Therefore, we recommend users be responsible and inclusive while using these generative models.
Note that our method only focuses on technical aspects. Both images and pre-trained models used in this paper will be open-released.

\end{document}